\documentclass{article}

    \PassOptionsToPackage{numbers, compress}{natbib}

 \usepackage[main, final]{neurips_2026}

\usepackage[utf8]{inputenc} 
\usepackage[T1]{fontenc}    
\usepackage{url}            
\usepackage{booktabs}       
\usepackage{amsfonts}       
\usepackage{nicefrac}       
\usepackage{microtype}      
\usepackage[table,dvipsnames]{xcolor}

\usepackage[
    colorlinks=true,
    citecolor=teal,
    linkcolor=purple,
    urlcolor=BlueViolet
]{hyperref}

\usepackage{booktabs}
\usepackage{adjustbox}
\usepackage{multirow}
\usepackage{bm}
\usepackage{makecell}

\usepackage{tabularx}
\usepackage{wrapfig}

\usepackage{tikz,graphicx,xcolor}
\usetikzlibrary{arrows.meta,positioning,fit,calc}
\usepackage{amssymb}
\usepackage{xspace}
\usepackage{caption}
\usepackage{amsmath}
\usepackage{soul}
\usepackage{subcaption}

\usepackage{hyperref}

\newcommand{\code}[1]{\texttt{#1}}
\usepackage{pifont}
\newcommand{\cmark}{\ding{51}}%
\newcommand{\xmark}{\ding{55}}%

\newcommand{\ourmethod}{{{TaxaAdapter}}\xspace}

\newcommand\mypara[1]{\noindent\textbf{#1}}

\newcommand{\wrt}{w.r.t.\xspace}
\newcommand{\etc}{etc.\xspace}

\newcommand{\treeoflife}{{\sc {TreeOfLife}}\xspace}
\newcommand{\flux}{{{FLUX.1-dev}}\xspace}

\newcommand{\gradientline}{
    \begin{center}
        \begin{tikzpicture}
            \shade[left color=gray!10, right color=gray!10, middle color=gray!50] 
                (0,0) rectangle (6cm, 0.2mm);
        \end{tikzpicture}
    \end{center}
}

\newcommand{\gradientseparator}{%
  \begin{tikzpicture}
    \shade[left color=gray!10, right color=gray!70]
      (0,0) rectangle (\linewidth,0.8pt);
  \end{tikzpicture}%
}
\usepackage{tcolorbox}

\usepackage{tocloft}

\newcommand\extrafootertext[1]{%
    \bgroup
    \renewcommand\thefootnote{\fnsymbol{footnote}}%
    \renewcommand\thempfootnote{\fnsymbol{mpfootnote}}%
    \footnotetext{#1}%
    \egroup
}



\newcommand{\subautoref}[2]{\hyperref[#1]{\autoref*{#1}(#2)}}

\title{TaxaAdapter: Scaling Fine-grained Species Image Generation to the Tree of Life}

\author{
Mridul Khurana$^{1\dagger}$ \quad Amin Karimi Monsefi$^{2}$ \quad Justin Lee$^{2}$ \quad Medha Sawhney$^{1}$ \\
{\bf  David Carlyn$^{2}$ \quad Julia Chae$^{3}$ \quad Jianyang Gu$^{2}$ \quad Rajiv Ramnath$^{2}$ }\\
{\bf Sara Beery$^{3}$ \quad Wei-Lun Chao$^{4}$ \quad Anuj Karpatne$^{5\dagger}$ \quad Cheng Zhang$^{6\dagger}$}
\\
\\
$^{1}$Virginia Tech \quad
$^{2}$The Ohio State University \quad
$^{3}$MIT \quad \\
$^{4}$Boston University \quad
$^{5}$University of Florida \quad
$^{6}$Texas A\&M University
\\
\\
\href{https://imageomics.github.io/TaxaAdapter/}{imageomics.github.io/TaxaAdapter}
}

\begin{document}

\maketitle

\extrafootertext{
$^\dagger$\texttt{\{mridul@vt.edu, akarptne@ufl.edu, chzhang@tamu.edu\}}.
}

\vspace{-15pt}
\begin{figure}[!h]
\centering
\includegraphics[width=\textwidth]{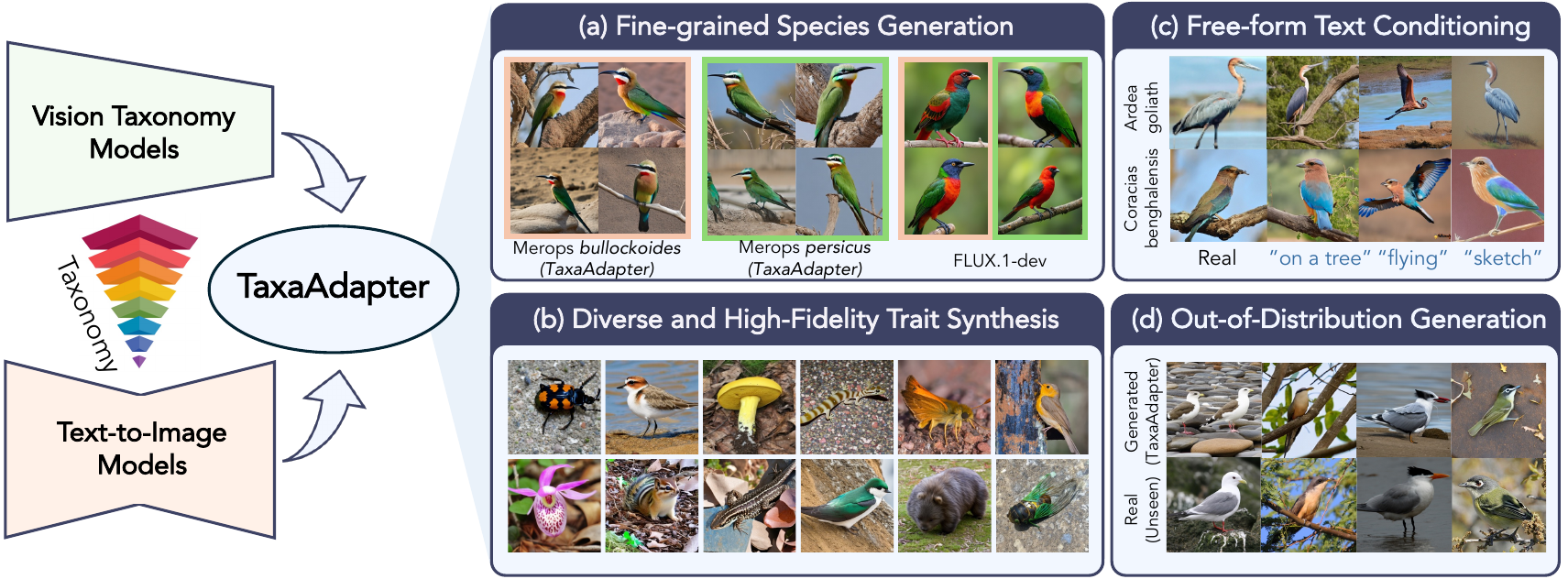}
\caption{
\small Existing text-to-image models struggle with fine-grained species details. 
We propose \textbf{\ourmethod} to inject VTM-derived embeddings into a frozen diffusion model to make fine-grained synthesis taxonomy-aware, enabling: (a) fine-grained species synthesis with accurate morphological details, (b) diverse and high-fidelity trait synthesis, (c) free-form text control, and (d) strong out-of-distribution generalization.}
\label{fig:teaser}
\vspace{-2mm}
\end{figure}
\begin{abstract}
Accurately generating images across the Tree of Life is difficult: there are over 10M distinct species on Earth, many of which differ only by subtle visual traits. Despite the remarkable progress in text-to-image synthesis, existing models often fail to capture the fine-grained visual cues that define species identity, even when their outputs appear photorealistic. 
To this end, we propose \ourmethod, a simple and lightweight adapter that incorporates the taxonomic embeddings of Vision Taxonomy Models (VTMs) such as BioCLIP to perform scalable fine-grained species generation over the entire Tree of Life.
\ourmethod injects VTM embeddings using taxonomy-text dual conditioning into a frozen text-to-image diffusion model, improving species-level fidelity while preserving flexible text control over attributes such as pose, style, and background.
Extensive experiments demonstrate that \ourmethod consistently improves morphology fidelity and species-identity accuracy over strong baselines.
To better evaluate these improvements, we also introduce a multimodal Large Language Model-based metric that summarizes trait-level descriptions from generated and real images, providing a more interpretable measure of morphological consistency.
Beyond showing improvements over standard benchmarking experiments, we observe that \ourmethod exhibits strong generalization capabilities in open-world settings, enabling species synthesis in challenging regimes such as few-shot species with only a handful of training images and even species unseen during training.
Overall, our results highlight that VTMs are a key ingredient for scalable, fine-grained species generation.

\end{abstract}

\section{Introduction}
\label{sec:intro}

Generating images of fine-grained species across the Tree of Life is a problem of growing importance in biodiversity science that presents a unique challenge for text-to-image synthesis. Unlike broad object categories, species identity is often defined by subtle morphological cues~\cite{van2017devil} such as slight differences in beak curvature or plumage patterns. For example, closely related bird species may differ only in faint head or throat markings  (Figure~\ref{fig:teaser}-(a)). In addition to this challenge, the underlying distribution of biodiversity images is extremely long-tailed: among the more than $10$M species on Earth, many have only a handful of curated photos. 
This motivates new methods for fine-grained generation that enable biologists to generate faithful visualizations of target species even in an open-world setting with little to no data, under varying contexts (e.g., pose) and rendering styles (e.g., schematic sketches) to discover diagnostic traits of closely related taxa.

Recent advances in text-to-image (T2I) diffusion models~\cite{rombach2022high, podell2023sdxl, esser2024scaling, flux2024, chen2023pixartalpha,zhang2023adding} trained on large-scale web data have reshaped image synthesis by strengthening the alignment between textual descriptions and visual details. 
However, these models predominantly learn representations organized around coarse semantic categories (e.g., ``bird'' or ``fish'') rather than precise taxonomic distinctions (e.g., \textit{Merops bullockoides}). Hence, they often struggle to capture the subtle morphological cues of individual species, producing images that appear realistic but belong to the wrong species (see example images from \flux \cite{flux2024} in Figure~\ref{fig:teaser}-(a)).
There is a growing body of work on learning taxonomic embeddings in diffusion models to improve fine-grained generation~\cite{monsefi2025taxadiffusion, khurana2024hierarchical, pan2024finediffusion}. 
Taxonomy hierarchically organizes species into nested groups based on shared traits, relationships, and evolutionary origins, providing a structured prior for species image generation.
While promising, these approaches require training these embeddings from scratch, making them computationally expensive to scale and limiting their ability to operate in unseen-species settings, hindering \textit{out-of-distribution} generation performance.

In parallel, there have been several recent efforts for learning taxonomy embeddings over the entire Tree of Life that are aligned with image embeddings for open-world recognition of species. This includes foundational works in building Vision Taxonomy Models (VTMs) such as BioCLIP~\cite{stevens2024bioclip,gu2025bioclip}, TaxaBind~\cite{sastry2025taxabind}, and BioTrove-CLIP~\cite{yang2024biotrove}. A unique feature of VTMs is that their taxonomy embeddings are pretrained to be \textit{aligned} with image-based features, however for a different goal of classification rather than image generation.
We ask the question: \textit{How can we leverage the image-aligned taxonomic embeddings of VTMs for open-world fine-grained image generation?}

We present \textbf{\ourmethod}, a lightweight adapter plug-in that injects taxonomy embeddings from VTMs into a frozen text-to-image diffusion model. We employ a simple dual conditioning design using text and taxonomic embeddings, where both these embeddings are sourced from frozen text encoders and taxonomy encoders of VTMs, respectively. This simplicity of our design---without specialized architectural changes or custom loss inventions---is the key strength of our work.
By aligning the conditioning space of generative models with image–aligned taxonomy embeddings, \ourmethod 
enables scalable and accurate fine-grained species generation (Figure~\ref{fig:teaser}-(b)) over the entire Tree of Life while preserving flexible attribute control through text prompts (Figure~\ref{fig:teaser}-(c)). 

Extensive experiments show that \ourmethod significantly improves species-level fidelity and morphological consistency compared with strong baselines. To better evaluate these improvements, we introduce a novel Multimodal Large Language Model (MLLM)-based trait evaluation framework that summarizes trait-level descriptions from generated and real images, providing an interpretable measure for morphological fidelity. 
Beyond these improvements on standard benchmark experiments, we show that \ourmethod exhibits strong generalization capabilities in out-of-distribution settings, enabling synthesis for rare species with limited training images as well as species never seen during training. 
To the best of our knowledge, \ourmethod is the first framework that bridges diffusion models and pre-trained VTMs, closing the gap between taxonomic knowledge and generative modeling, and offering a practical route for scaling fine-grained generation over the Tree of Life.

\section{Related Work}
\label{sec:related_works}

\mypara{Fine-grained Generation}
aims to model categories with subtle visual differences, such as closely related fish species, fine facial attributes, or detailed object variations. Prior work explores several directions. Sketch-based control methods guide synthesis through user-provided structure, enabling local detail refinement~\cite{navard2024knobgen, zhao2023uni, li2025controlnet, zhang2023adding}. Attribute-manipulation approaches condition on latent or semantic attributes to adjust specific features~\cite{huang2024diffusion, kim2022diffusionclip, yue2023chatface, han2023highly, matsunaga2022fine,zhang2023iti}. Personalized generation aims to preserve identity or instance-specific features by tuning part of the model~\cite{zhang2024survey, ruiz2023dreambooth, li2024stylegan, wang2024animatelcm, ruiz2024hyperdreambooth}. 
Recently, TaxaDiffusion~\cite{monsefi2025taxadiffusion} and PhyloDiffusion~\cite{khurana2024hierarchical} explore conditioning on hierarchical knowledge (e.g., taxonomy) for species image synthesis. 
However, training taxonomy embeddings in these models from scratch is expensive and limits their scope only to species seen during training. Moreover, retraining the embedding space of diffusion models risks overwriting its native text-conditioning pathway, limiting their flexibility to be guided with text prompts.

\mypara{Conditional Image Generation}
extends diffusion models with additional signals such as segmentation masks, sketches, depth maps, or learned adapters~\cite{pan2024finediffusion, li2025controlnet, zhang2023adding}. These methods control structure or appearance using auxiliary inputs or small parameter updates, often through cross-attention or feature modulation.
However, existing conditioning methods assume that the conditioning input already carries meaningful visual information. Taxonomic text does not meet this requirement: it encodes relationships, not appearance. As a result, zero-shot conditioning on taxonomic names provides limited benefit in fine-grained biological settings. Another approach for fine-grained generation is in-context learning by prompting example images of species  \cite{labs2025flux}. However, this violates the open-world setting where we want to generate images even for novel or unseen species.

\mypara{Vision Taxonomy Models}~\cite{stevens2024bioclip, gu2025bioclip,sastry2025taxabind,yang2024biotrove}
align images and taxonomy in a shared embedding space. CLIP~\cite{radford2021learning} and its variants learn general-purpose aligned embeddings, while domain-specific models such as BioCLIP~\cite{stevens2024bioclip, gu2025bioclip}, TaxaBind~\cite{sastry2025taxabind} \& BioTrove-CLIP~\cite{yang2024biotrove} extend this idea by pairing species images with taxonomic information. These models organize taxa according to visual similarity and have shown promise for classification and retrieval tasks.

\section{\ourmethod}

\ourmethod solves the twin goals of fine-grained generation (preserving morphological identity of species) and open-world generation (generalizing over the Tree of Life), making it different from all previous works in this direction. We first provide a motivation for why species names are not sufficient for fine-grained generation and why injecting taxonomic embeddings is non-trivial for the entire Tree of Life. We then introduce our key innovation of taxonomy-text dual conditioning.

\subsection{Why Species Names are Not Sufficient for Fine-grained Generation?}
Taxonomic names of species alone provide weak visual conditions for text encoders such as CLIP~\cite{radford2021learning} or T5~\cite{raffel2020exploring}, as they are trained largely for broad semantic alignment rather than fine-grained biological morphology. For example, a prompt such as ``a photo of \textit{Merops persicus}'' identifies a species taxonomically, but does not explicitly describe the visual traits such as wing patterns, beak shape, or body proportions that uniquely differentiate a species, creating a mismatch between taxonomic names and image features. As a result, even strong T2I models can fail to preserve the morphological identity of species when conditioned solely on species names.
Note that this is different from the case where explicit visual attributes are supplied in the text prompt, such as ``a photorealistic painting of a bird with \textit{green} wings sitting on a tree'', which T2I models are trained to handle well. We thus need to move beyond species names to taxonomic embeddings that are aligned with visual information.

\begin{figure}[t]
\centering
\includegraphics[width=0.9\textwidth]{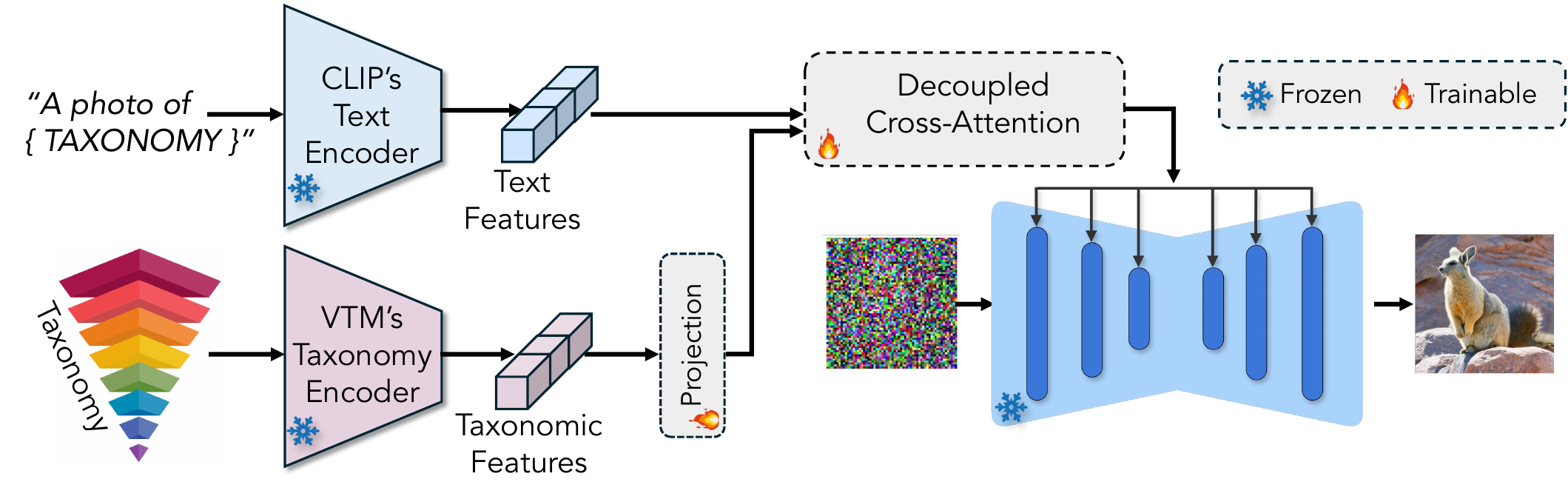}
    \caption{\small \ourmethod pipeline overview. Given a taxonomic name (e.g., \code{Kingdom} $\rightarrow$ \code{Species}), we extract taxonomy-image aligned embeddings using a frozen vision taxonomy model (e.g., BioCLIP~\cite{gu2025bioclip}) and obtain complementary text features from the frozen CLIP text encoder. The dual conditioning streams are fused through a decoupled cross-attention mechanism, while the diffusion backbone remains frozen.
    }
    \label{fig:framework}
\vspace{-4mm}
\end{figure}

\subsection{Why is it Non-trivial to Inject Taxonomic Embeddings in Diffusion Models?}
\label{sec:non-trivial}
A natural way to incorporate taxonomy in T2I models is to fine-tune taxonomic embeddings over a known set of species.
However, fine-tuning the backbone (or LoRA) to map thousands of taxonomic names to morphology from scratch, even with progressive training \cite{monsefi2025taxadiffusion}, does not scale well in long-tailed datasets where many species have only a few images, and can degrade the model’s original text control.
The second natural way to leverage taxonomic embeddings from VTMs for open-world generation is to directly swap the text encoder in T2I models with VTM's taxonomic encoder. However, since diffusion backbones are pretrained for text inputs and not VTM embeddings, there is an interface mismatch that leads to inferior generation quality, as shown empirically in Section~\ref{sec:ablations}. 

\subsection{Proposed Idea: Taxonomy-Text Dual Conditioning}
\label{sec:idea}

\ourmethod adopts a dual conditioning mechanism to preserve the ability of T2I models to be controlled using free-form text prompts while being grounded in image-aligned taxonomic embeddings provided by pre-trained VTMs. Specifically, \ourmethod employs an adapter-style~\cite{ye2023ip, mou2024t2i,li2025controlnet}, decoupled conditioning design inspired by prior works on conditional diffusion models~\cite{zhang2023adding,ye2023ip,li2024photomaker,zhao2023uni}. 
We leave the original text-conditioning pathway untouched and attach a separate, lightweight branch that carries the VTM taxonomy embedding into the U-Net through an additional cross-attention module, as illustrated in Figure~\ref{fig:framework} (see Section~\ref{subsec:setup} and Section~\ref{subsec:background} for implementation and training details).
The VTM branch is responsible for injecting taxonomy–image aligned information learned from biological pretraining, while the text branch preserves the pretrained model’s ability to follow natural-language prompts across arbitrary contexts.
A key realization of our work is that instead of introducing a new architectural primitive, we can \textit{repurpose adapter-style training for scaling diffusion models to the entire Tree-of-Life while being morphologically accurate}.

\mypara{Why is \ourmethod Novel?} 
Replacing the image conditioning pathway of adapter-style methods with taxonomic embeddings is not an obvious choice, since it is more natural to treat taxonomy as text. Indeed, previous works in fine-grained generation such as TaxaDiffusion \cite{monsefi2025taxadiffusion}  
use taxonomic names as inputs in the text encoder of CLIP. However, taxonomic names alone do not provide visual information necessary for fine-grained generation. This is where we leverage the
unique feature of VTM taxonomic embeddings that are pretrained to be \textit{aligned with image-based features} as a complementary source of visual information in the dual conditioning mechanism of \ourmethod. Conditioning on VTM taxonomic embeddings also enables high-level control of morphological identity across all images of a species rather than low-level spatial control over a particular image instance using sketches or segmentation maps, as explored in prior adapter-style works \cite{ye2023ip, mou2024t2i,li2025controlnet}.

\begin{figure}[h]
    \centering
    \includegraphics[width=1.0\linewidth]{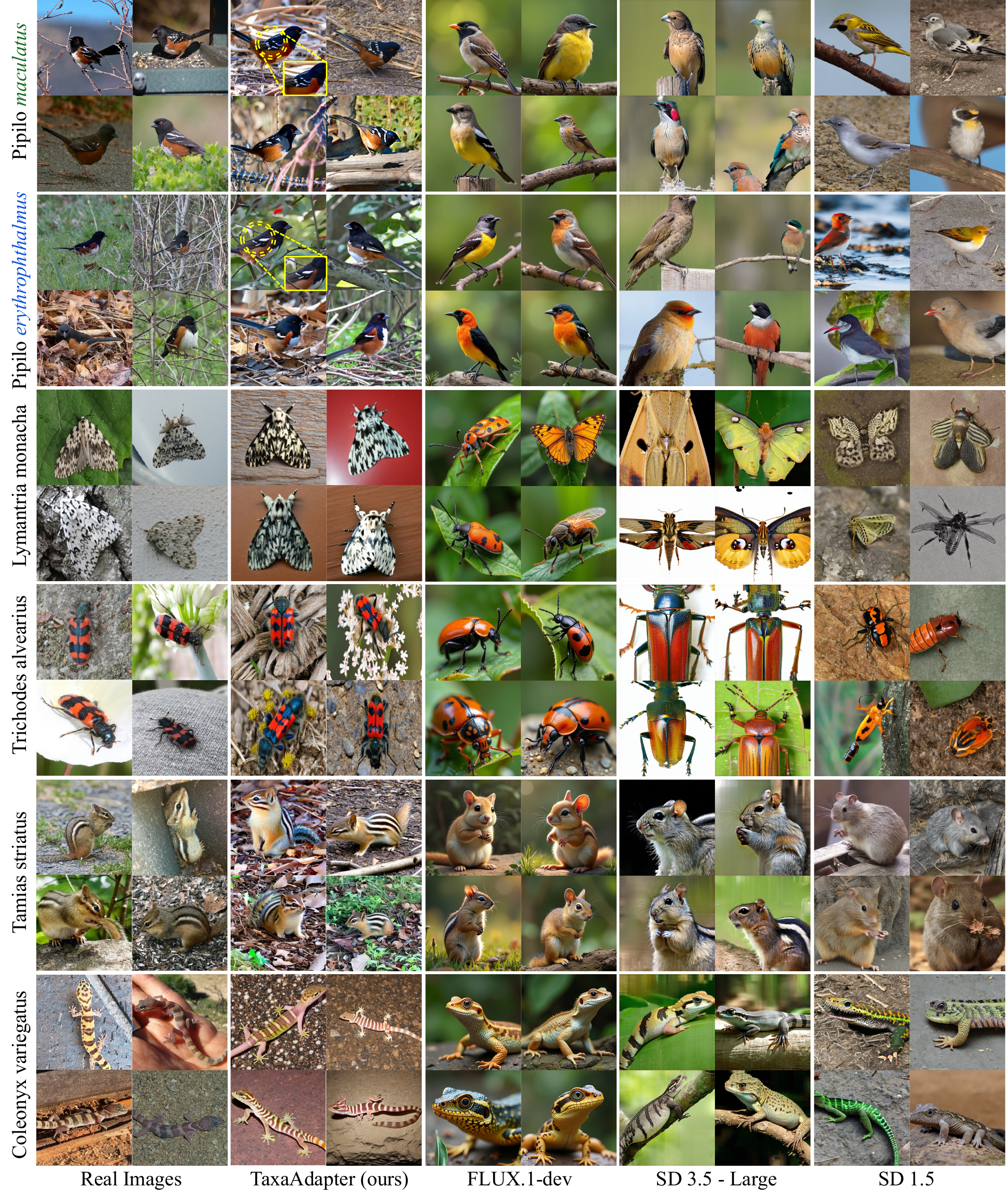}
    \caption{\small Qualitative comparison on \treeoflife-1M. Each row shows a different species spanning birds, mammals, insects, and reptiles. \ourmethod generates morphology-faithful images that align with taxonomy-defining traits (e.g., texture patterns, body shape, coloration) while maintaining realistic textures and backgrounds. 
    Notably, the first two rows illustrate two \textbf{extremely similar species} under the same \code{Genus}, where subtle differences in white spotting patterns on bird wings (highlighted in yellow circles) are correctly captured. Please see Appendix~\ref{sup:additional_qualitative} for additional qualitative results.}
    \label{fig:qual_tol}
    \vspace{-8pt}
\end{figure}

\mypara{What Does \ourmethod Enable?}
Our work, for the first time, enables morphologically accurate fine-grained species generation over the entire Tree of Life, even for species never seen during training. 
Because VTMs organize taxa according to visual and taxonomic structure, a species with limited training images can still benefit from representations learned from related taxa at the \code{Genus}, \code{Family}, or higher taxonomic levels, enabling open-world generation.
The first two rows of Fig.~\ref{fig:qual_tol} show \textit{extremely similar} species that belong to the same genus (\textit{Pipilo}). Their subtle trait variations on wing color patterns (see yellow-highlighted boxes) are accurately captured in images generated by \ourmethod, demonstrating strong semantic alignment in the taxonomy–image space.
On the other hand, images from T2I models appear photorealistic but are not faithful to species identity. 
\ourmethod also preserves free-form text prompting while maintaining taxonomic identity.

\section{Proposed Framework for Trait Fidelity Evaluation}
\label{subsec:vlm_metric}

\begin{figure}[t]
\centering
\includegraphics[width=1\textwidth]{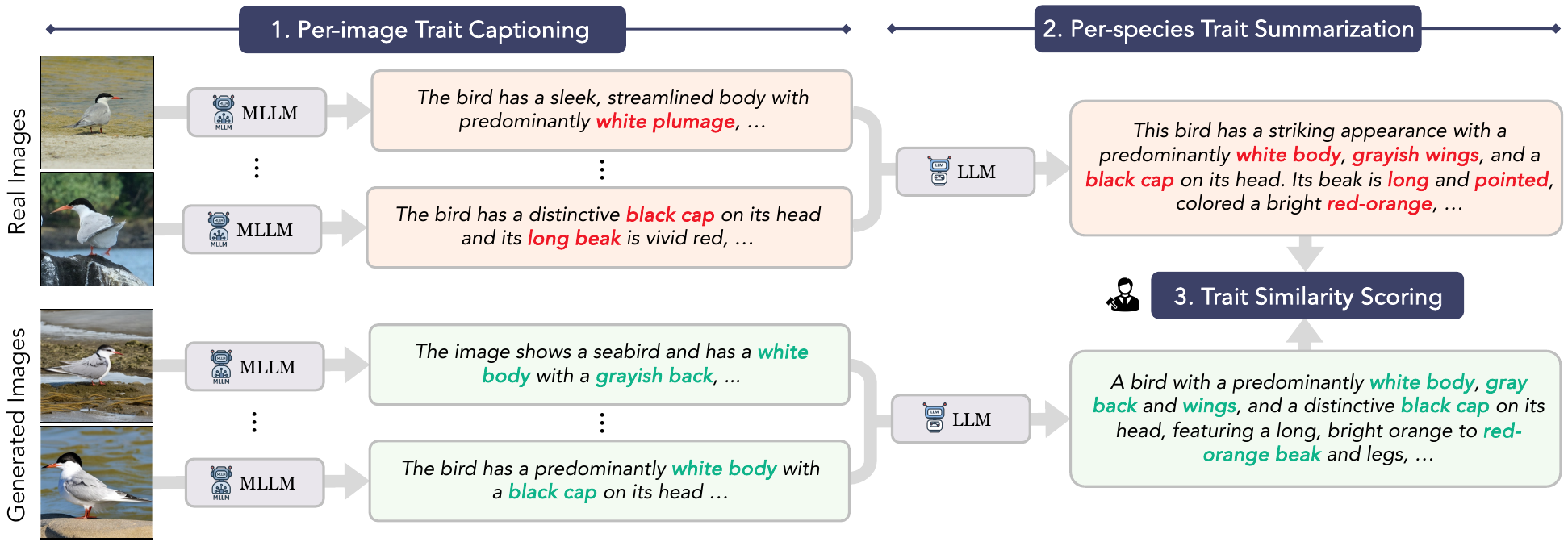}
    \caption{\small Caption-based trait fidelity evaluation. We leverage an MLLM to generate trait captions for real and generated images, summarize each set into a species-level trait description with an LLM, and compute text similarity between the two summaries. Our metric provides an interpretable, trait-level measure of morphological fidelity that complements standard image metrics.}
    \label{fig:vlm_metric}
\vspace{-2mm}
\end{figure}

Evaluating fine-grained species generation is challenging since standard metrics such as FID~\cite{Seitzer2020FID} and LPIPS~\cite{zhang2018unreasonable} emphasize low-level structure (e.g., color and texture) but correlate weakly with species-defining morphological traits. Expert-annotated, trait-level descriptions for species are difficult to obtain at scale. Therefore, we propose a scalable trait-fidelity metric based on trait captions generated by MLLMs. We generated captions of both real and generated images and use the real image captions as a reference. 
Figure~\ref{fig:vlm_metric} illustrates the pipeline, which consists of three steps:

\mypara{Step 1: Per-image Trait Captioning.} For each species, we prompt an MLLM (namely, {InternVL3-8B~\cite{zhu2025internvl3}}) to describe only visible traits such as shape, color, and texture, while explicitly excluding species names to avoid bias (see Appendix~\ref{sup:trait_fidelity_appendix} for prompt details). In our experiments, we sample 10 images per species for the generated set, resulting in 10 corresponding trait captions. For real images, we caption all the available images of that species.

\mypara{Step 2: Per-species Trait Summarization.} Next, we employ an LLM (namely {LLaMa3.1-8B-Instruct~\cite{grattafiori2024llama}}) to consolidate the per-image captions into a single ``trait caption'' for each species. We instruct the model to extract and merge only morphological attributes (e.g., ``yellow crown'', ``short beak'', ``dark wings'') while discarding contextual elements such as background or lighting (see Appendix~\ref{sup:trait_fidelity_appendix} for prompt details). The same summarization procedure is applied to real images, producing one reference trait caption per species. Our key insight is that per-species summarization helps recover the full set of visual traits, since distinctive features may be occluded or absent in individual species images due to variations in pose, orientation, or visibility.
    
\mypara{Step 3: Trait Similarity Scoring}. Finally, we measure the similarity between the summarized captions of real and generated images using several text-based metrics, including BERTScore~\cite{zhang2019bertscore}, CLIPScore~\cite{hessel2021clipscore}, ROUGE~\cite{lin-2004-rouge}, and Mistral-7B text similarity~\cite{jiang2023mistral7b}, providing a quantitative assessment of how faithfully the generated species preserve authentic visual traits.

\mypara{Remark.}
Since expert trait annotations are difficult to obtain at scale, we use real images as references and apply the same captioning and summarization pipeline to both real and generated images, which helps reduce systematic bias and makes the evaluation scalable.
We further validate this protocol through a user study (Appendix~\ref{sup:user_study}) and by comparing against BioCAP~\cite{zhang2025biocap} reference captions, which provide Wikipedia-grounded, trait-rich descriptions (Appendix~\ref{sup:biocap_grounding}). Additional implementation details are provided in Appendix~\ref{sup:impl}.
\section{Experiments}
\label{sec:experiments}

\subsection{Experimental Setup}
\label{subsec:setup}

\mypara{Implementation Details.}
We repurpose the IP-Adapter~\cite{ye2023ip} training paradigm for taxonomy-text dual conditioning. We replace the image prompt encoder with VTM's taxonomic encoder and pass the full taxonomic hierarchy from \code{Kingdom} to \code{Species} to obtain the taxonomic embeddings. These embeddings are projected into the diffusion model's conditioning interface and injected through the decoupled cross-attention as shown in Figure~\ref{fig:framework}, while keeping the VTM encoder, original text encoder, and diffusion backbone frozen. We consider three VTM backbones: BioCLIP~\cite{gu2025bioclip}, BioTrove-CLIP~\cite{yang2024biotrove}, and TaxaBind~\cite{sastry2025taxabind} and Stable Diffusion v1.5~\cite{rombach2022high} as the frozen diffusion backbone. Additional results with SDXL backbone are provided in Appendix~\ref{sup:sdxl}.

\mypara{Training Details.}
We train our model for 100 epochs with a batch size of 64 per GPU. We use AdamW with a learning rate of $1\times10^{-4}$, weight decay of $0.01$, and mixed-precision training. At inference, we generate images at $512 \times 512$ resolution using 50 denoising steps. We set the taxonomy-conditioning scale to $\lambda=1$ for all quantitative experiments and vary it only in the free-form text-control analysis. Additional experiments with varying hyperparameters are provided in Appendix~\ref{sup:impl}.

\mypara{Baselines.}
We use pre-trained text-to-image models including SD v1.5~\cite{rombach2022high}, SD v3.5-Large~\cite{esser2024scaling}, and FLUX.1-dev~\cite{flux2024} as baselines. For every model, we either apply them zero-shot (ZS) or perform LoRA fine-tuning~\cite{hu2022lora}. We also consider a state-of-the-art taxonomy-aware generative model, TaxaDiffusion~\cite{monsefi2025taxadiffusion} as a baseline. See Appendix~\ref{sup:impl} for additional details about baselines.

\mypara{Datasets.}
We consider three biodiversity datasets for training:
(1)~iNat-mini~\cite{van2018inaturalist}, with 500K images spanning 10K species from plants, animals, and fungi (covering all seven taxonomic levels), 
(2)~\treeoflife-1M, a subset of \treeoflife-10M dataset~\cite{stevens2024bioclip} with 1M images spanning 200K+ species, and
(3)~FishNet~\cite{khan2023fishnet}, with 95K images spanning 17.4K fish species.
For every image in these datasets, we use their full taxonomic hierarchy string (\code{Kingdom} $\rightarrow$ \code{Species}).

\mypara{Metrics.}
We evaluate all models across five criteria:
~\textbf{(1) Image Quality:} We use standard metrics of Fréchet Inception Distance (FID)~\cite{Seitzer2020FID} and Learned Perceptual Image Patch Similarity (LPIPS)~\cite{zhang2018unreasonable}.
~\textbf{(2) Classification Accuracy:} To evaluate whether generated images preserve species-level identity, we report top-$k$ Classification Accuracy Score (CAS@k)~\cite{ravuri2019classification} using a DINOv2~\cite{oquab2023dinov2} linear-probe classifier trained on a separate 500K subset of real images taken from remaining iNaturalist~\cite{van2018inaturalist}/ {\sc{ToL-10M}}~\cite{stevens2024bioclip}.
~\textbf{(3) Text-to-Image Alignment:} 
We report CLIPScore~\cite{hessel2021clipscore} to measure alignment between generated images and input prompts. 
\textbf{(4) Diversity}: We report Precision and Recall~\cite{kynkaanniemi2019improved} to assess fidelity and coverage. 
~\textbf{(5) Caption-based Trait Fidelity:} We use our proposed MLLM-driven trait similarity pipeline from Section~\ref{subsec:vlm_metric} to measure morphological trait fidelity in generated images.
See Appendix~\ref{sup:impl} and \ref{sup:main_results} for implementation details of these metrics.

\mypara{Remark.}
While stronger backbones such as SD3.5-Large and FLUX.1-dev improve open-domain generation, they still struggle with fine-grained species identity under generic text prompting. This suggests that scaling the generator alone is insufficient for biodiversity synthesis.
We use SD1.5 as the primary backbone, as it allows lightweight training of VTM conditioning (22M params) while keeping the T2I model frozen. Applying the same adapter design to large diffusion-transformer backbones, such as FLUX.1-dev, would require substantially more compute (500M params~\cite{xlabs_flux_ip_adapter}).

\subsection{Results}

\begin{table}[t]
\centering
\tabcolsep 1pt
\small
\caption{\small Results on iNat-mini over 1,103 species (one from each \code{Family} level in the taxonomy).
ZS denotes zero-shot results. We show \ourmethod{} results using two VTM text encoders, BioCLIP and TaxaBind. 
\textbf{Best} is bolded, \underline{second-best} underlined.
}
\resizebox{\columnwidth}{!}{
\begin{tabular}{lccccccccccc}
\toprule
\multirow{2.5}{*}{Model} 
& \multicolumn{2}{c}{Image Quality} 
& \multicolumn{2}{c}{Classification Accuracy} 
& \multicolumn{1}{c}{T2I Alignment} 
& \multicolumn{2}{c}{Diversity}
& \multicolumn{4}{c}{Caption-based Trait Fidelity}  \\
\cmidrule(lr){2-3} 
\cmidrule(lr){4-5} 
\cmidrule(lr){6-6} 
\cmidrule(lr){7-8}
\cmidrule(lr){9-12}
 & {FID$\downarrow$} 
 & {LPIPS$\downarrow$} 
 & {CAS@1$\uparrow$} 
 & {CAS@5$\uparrow$} 
 & {CLIPScore$\uparrow$} 
 & {Prec.$\uparrow$} 
 & {Rec.$\uparrow$} 
 & {BERT$\uparrow$} 
 & {CLIP $\uparrow$} 
 & {ROUGE-L $\uparrow$} 
 & {Mistral-7B $\uparrow$} \\
\midrule

{SD 1.5 (ZS)~\cite{rombach2022high}}        
& 31.43 & 0.768 & 1.24\% & 2.86\% & 18.96 & 0.45 & 0.32 & 0.22 & 0.77 & 0.20 & 0.89 \\

{SD 3.5-Large (ZS)~\cite{esser2024scaling}} 
& 70.63 & 0.806 & 1.86\% & 4.48\% & 19.46 & 0.38 & 0.15 & 0.23 & 0.78 & 0.21 & 0.89 \\

{\flux (ZS)~\cite{flux2024}}                
& 74.14 & 0.801 & 1.78\% & 4.21\% & 18.94 & 0.55 & 0.05 & 0.29 & 0.79 & \underline{0.24} & 0.91 \\

\midrule

SD 1.5 (LoRA~\cite{hu2022lora})       
& 15.14 & \textbf{0.724} & 30.69\% & 56.01\% & 22.84 & \underline{0.66} & 0.51 & 0.32 & 0.80 & 0.25 & \underline{0.92} \\

SD 3.5-Large (LoRA~\cite{hu2022lora})       
& 31.42 & 0.762 & 3.47\% & 10.18\% & 21.69 & 0.62 & 0.28 & 0.29 & 0.79 & \underline{0.24} & 0.91 \\

\flux (LoRA~\cite{hu2022lora})              
& 49.72 & 0.765 & 1.81\% & 5.10\% & 20.06 & 0.57 & 0.26 & 0.26 & 0.77 & 0.23 & 0.90 \\
\midrule

\textbf{\ourmethod~(w/BioCLIP)}             
& \underline{13.31} & 0.742 & \textbf{61.36\%} & \textbf{81.99\%} & \textbf{23.03} & \textbf{0.71} & \underline{0.52} & \textbf{0.33} & \textbf{0.82} & \textbf{0.26} & \textbf{0.93} \\

\textbf{\ourmethod~(w/TaxaBind)}            
& \textbf{13.04} & \underline{0.741} & \underline{59.98\%} & \underline{79.85\%} & \underline{22.89} & \textbf{0.71} & \textbf{0.53} & \underline{0.32} & \underline{0.81} & \textbf{0.26} & \underline{0.92} \\

\bottomrule
\end{tabular}
}
\vspace{-3mm}
\label{tab:inat}
\end{table}

\begin{table}[t]
\centering
\tabcolsep 1pt
\small
\caption{\small Results on \treeoflife-1M over 500 species. 
}
\resizebox{\columnwidth}{!}{
\begin{tabular}{lccccccccccc}
\toprule
\multirow{2.5}{*}{Model} 
& \multicolumn{2}{c}{Image Quality} 
& \multicolumn{2}{c}{Classification Accuracy} 
& \multicolumn{1}{c}{T2I Alignment} 
& \multicolumn{2}{c}{Diversity} 
& \multicolumn{4}{c}{Caption-based Trait Fidelity}  \\
\cmidrule(lr){2-3} 
\cmidrule(lr){4-5} 
\cmidrule(lr){6-6} 
\cmidrule(lr){7-8} 
\cmidrule(lr){9-12}
 & {FID$\downarrow$} 
 & {LPIPS$\downarrow$} 
 & {CAS@1$\uparrow$} 
 & {CAS@5$\uparrow$} 
 & {CLIPScore$\uparrow$} 
 & {Prec.$\uparrow$} 
 & {Rec.$\uparrow$} 
 & {BERT$\uparrow$} 
 & {CLIP $\uparrow$} 
 & {ROUGE-L $\uparrow$} 
 & {Mistral-7B $\uparrow$} \\
\midrule
{SD 1.5 (ZS)~\cite{rombach2022high}}        
& {46.89} & 0.7973 & 9.32\%  & 25.54\%  & 19.19 & 0.54 & 0.33 & 0.28 & 0.75 & 0.24 & 0.90   \\

{SD 3.5-Large (ZS)~\cite{esser2024scaling}} 
& 79.26 & 0.8179 & 11.00\% & 29.32\%  & 19.43 & 0.55 & 0.18 & 0.28 & 0.74 & 0.24 & 0.90   \\

{\flux (ZS)~\cite{flux2024}}                
& 90.09 & 0.8003 & 12.44\% & 31.14\%  & 19.00 & 0.66 & 0.06 & \underline{0.30} & 0.75 & \underline{0.25} & 0.90  \\
\midrule

SD 3.5-Large (LoRA~\cite{hu2022lora})  
& 64.21 & 0.7977 & {20.46}\% & {50.36}\%  & \underline{20.97} & 0.46 & 0.28 & 0.28 & \underline{0.78} & 0.24 & \underline{0.91}     \\

\flux (LoRA~\cite{hu2022lora})         
& 59.13 & {0.7703} & 13.06\% & 34.70\%  & 20.01 & \textbf{0.69} & 0.32 & 0.28 & 0.77 & \underline{0.25} & \underline{0.91}    \\
\midrule

\textbf{\ourmethod~(w/BioCLIP)}  
& \textbf{29.87} & \textbf{0.7345} & \textbf{61.98\%} & \textbf{82.96\%} & \textbf{21.76} & \underline{0.67} & \underline{0.51} & \textbf{0.33} & \textbf{0.81} & \textbf{0.28} & \textbf{0.92} \\

\textbf{\ourmethod~(w/TaxaBind)}  
& \underline{33.78} & \underline{0.7476} & \underline{52.22\%} & \underline{75.24\%} & 20.90 & 0.59 & \textbf{0.57} & 0.27 & 0.77 & \underline{0.25} & \underline{0.91} \\

\bottomrule
\end{tabular}
}
\vspace{-3mm}
\label{tab:tol-1m-1}
\end{table}


\begin{table*}[t]
\centering
\small
\setlength{\tabcolsep}{3pt}

\begin{minipage}[t]{0.48\textwidth}
\centering
\caption{\small Quantitative comparison with TaxaDiffusion~\cite{monsefi2025taxadiffusion} on 156 species of iNat-mini and 200 species of FishNet following the evaluation protocol of~\cite{monsefi2025taxadiffusion}. TaxaDiffusion numbers directly copied from \cite{monsefi2025taxadiffusion}.}
\vspace{2pt}
\resizebox{\linewidth}{!}{
\begin{tabular}{@{}lcccc@{}}
\toprule
\multirow{2}{*}{Model} 
& \multicolumn{2}{c}{\textbf{iNat-mini}} 
& \multicolumn{2}{c}{\textbf{FishNet}} \\
\cmidrule(lr){2-3} \cmidrule(lr){4-5}
& {FID$\downarrow$} & {LPIPS$\downarrow$} 
& {FID$\downarrow$} & {LPIPS$\downarrow$} \\
\midrule
SD 1.5 (ZS)~\cite{rombach2022high}      
& 66.91 & 0.779 
& 61.93 & 0.7737  \\
TaxaDiffusion~\cite{monsefi2025taxadiffusion}    
& 46.39 & 0.747 
& 31.87 & \textbf{0.7319} \\
\midrule
\textbf{\ourmethod} 
& \textbf{30.36} & \textbf{0.741} 
& \textbf{30.68} & 0.7818 \\
\bottomrule
\end{tabular}
}
\label{tab:fishnet}
\end{minipage}
\hfill
\begin{minipage}[t]{0.47\textwidth}
\centering
\caption{\small Long-tailed species generation results on ToL-1M. We report results for 5K species with only $1$ training image and fewer than $5$ training images.}
\vspace{2pt}
\resizebox{\linewidth}{!}{
\begin{tabular}{@{}lcccc@{}}
\toprule
\multirow{2}{*}{Model} 
& \multicolumn{2}{c}{\textbf{Single image}} 
& \multicolumn{2}{c}{\textbf{$<5$ images}} \\
\cmidrule(lr){2-3}\cmidrule(lr){4-5}
& {FID$\downarrow$} & {LPIPS$\downarrow$} 
& {FID$\downarrow$} & {LPIPS$\downarrow$}  \\
\midrule
SD 1.5~\cite{rombach2022high}        
& \underline{43.44} & \underline{0.797} 
& \underline{43.07} & \underline{0.799} \\
SD 3.5-Large~\cite{esser2024scaling}  
& 68.66 & 0.817 
& 71.42 & 0.819 \\
FLUX.1-dev~\cite{flux2024}   
& 89.87 & 0.821 
& 80.24 & 0.811\\
\midrule
\textbf{\ourmethod} 
& \textbf{32.19} & \textbf{0.733}  
& \textbf{35.43} & \textbf{0.724} \\
\bottomrule
\end{tabular}
}
\label{tab:long_tailed}
\end{minipage}
\end{table*}

We summarize results across the three benchmarks in Tables~\ref{tab:inat},~\ref{tab:tol-1m-1}, and~\ref{tab:fishnet}, where \ourmethod demonstrates consistent gains in image quality, taxonomy-alignment, species identity, and trait fidelity. 
While zero-shot T2I models generate visually realistic images (high precision for FLUX.1-dev), but they substantially underperform on species identity (CAS), failing to preserve fine-grained species identity. LoRA fine-tuning yields only marginal CAS improvements.
In contrast, \ourmethod achieves strong identity preservation (up to $\sim$83\%) while also improving the other metric categories.
Table~\ref{tab:fishnet} compares \ourmethod against TaxaDiffusion on the same subset of iNat-mini and FishNet species as reported in \cite{monsefi2025taxadiffusion}.
Figure~\ref{fig:qual_tol} provides visual comparisons across species drawn from multiple taxonomy groups. 
Appendix~\ref{sup:biocap_t2i} further shows that providing detailed trait descriptions of species appearance, traits, and morphology from BioCAP~\cite{zhang2025biocap} that are grounded in Wikipedia as input text to T2I models does improve the baseline generation; however, the models still fall behind \ourmethod.
Please see Appendix~\ref{sup:additional_results} for additional quantitative and qualitative results.

\subsection{Generalization Capabilities of \ourmethod{}}
\label{sec:emergent}

\mypara{Can \ourmethod Generate Rare Species?}
Biodiversity datasets are intrinsically long-tailed, with many species under-observed due to geography, seasonality, and sampling bias, resulting in very few images for a broad range of taxa. For biological use, a generative model should remain species-faithful in this \emph{rare-but-seen} regime, where only a handful of examples exist to anchor morphology. 
Table~\ref{tab:long_tailed} reports results on 5000 \treeoflife-1M species with only a single training image and with fewer than five training images, where we see \ourmethod{} outperform baselines. We do not report CAS here because reliable species classifiers cannot be trained for species with only one or a few images.

\begin{figure*}[h]
\centering

\begin{minipage}[t]{0.48\textwidth}
\vspace{0pt}
  \centering
  \captionof{table}{\small Quantitative OOD evaluation on 51 CUB-200-2011 species unseen during training. Models are trained on iNat-mini and tested on unseen CUB classes. 
  }
  \label{tab:cub-ood}
  \vspace{2mm}
  \resizebox{\linewidth}{!}{
    \begin{tabular}{lcccc}
      \toprule
      Model & FID $\downarrow$ & LPIPS $\downarrow$ & CAS@1 $\uparrow$ & CAS@5 $\uparrow$\\
      \midrule
      SD 1.5~\cite{rombach2022high} & \textbf{36.13} & \underline{0.7853} & \underline{9.61\%}  & \underline{25.88\%} \\
      SD 3.5-Large~\cite{esser2024scaling} & 44.21 & 0.8016 &  8.24\% & 23.92\% \\
      \flux~\cite{flux2024} & 60.32 & 0.7901 & 6.67\% & 20.39\%\\
      \midrule
      \textbf{\ourmethod} & \underline{42.36} & \textbf{0.7777} & \textbf{38.82\%}  & \textbf{68.24\%} \\
      \bottomrule
      \vspace{-8mm}
    \end{tabular}
    }
\end{minipage}\hfill
\begin{minipage}[t]{0.48\textwidth}
\vspace{0pt}
  \centering
  \includegraphics[width=0.9\linewidth]{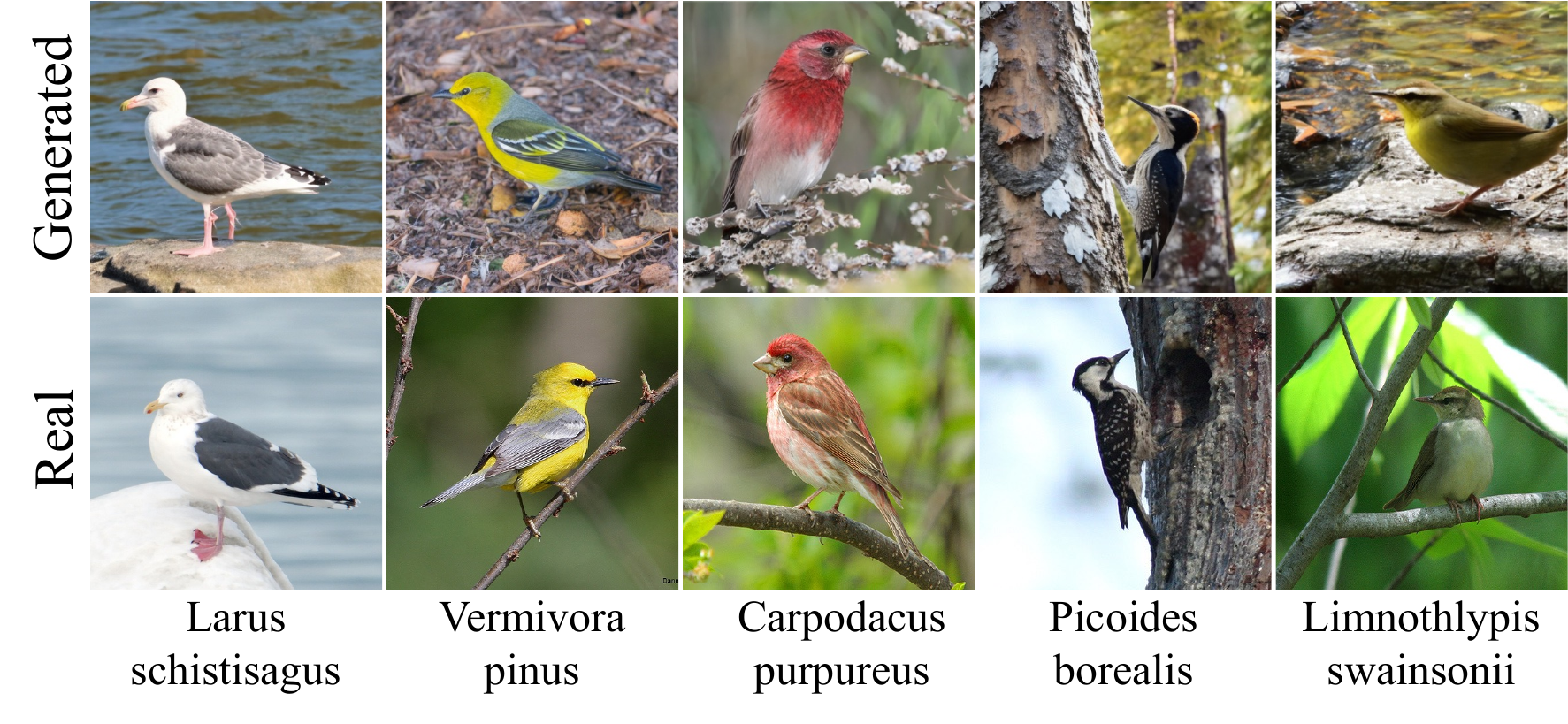}
  \vspace{-3pt}
  \captionof{figure}{\small Qualitative results on OOD species generation. \ourmethod demonstrates strong zero-shot morphological consistency on unseen classes.}
  \label{fig:ood}
\end{minipage}
\end{figure*}

\mypara{Can \ourmethod Synthesize Unseen Species?}
Many species remain under-documented in existing image datasets, and new observations continually arrive from the field.
Unlike long-tailed evaluation (rare but seen species), this setting probes whether \ourmethod enables accurate generation of species never seen during training. 
We identify 51 species from the CUB-200 dataset~\cite{WahCUB_200_2011} that have no overlap with iNat-mini and use models trained on iNat-mini to synthesize images for these unseen classes.
Figure~\ref{fig:ood} shows that \ourmethod generates images that closely resemble the real species despite never seeing them during training. Table~\ref{tab:cub-ood} also confirms this observation: \ourmethod achieves competitive similarity scores compared with zero-shot \flux and SD 3.5, demonstrating that coupling diffusion models with VTMs enables Out-of-Distribution (OOD) species generation. Appendix~\ref{sup:additional_ood_results} reports additional results on the FishNet~\cite{khan2023fishnet} dataset and separates the species seen during VTM pretraining.


\begin{figure}[t]
\centering
\includegraphics[width=\linewidth]{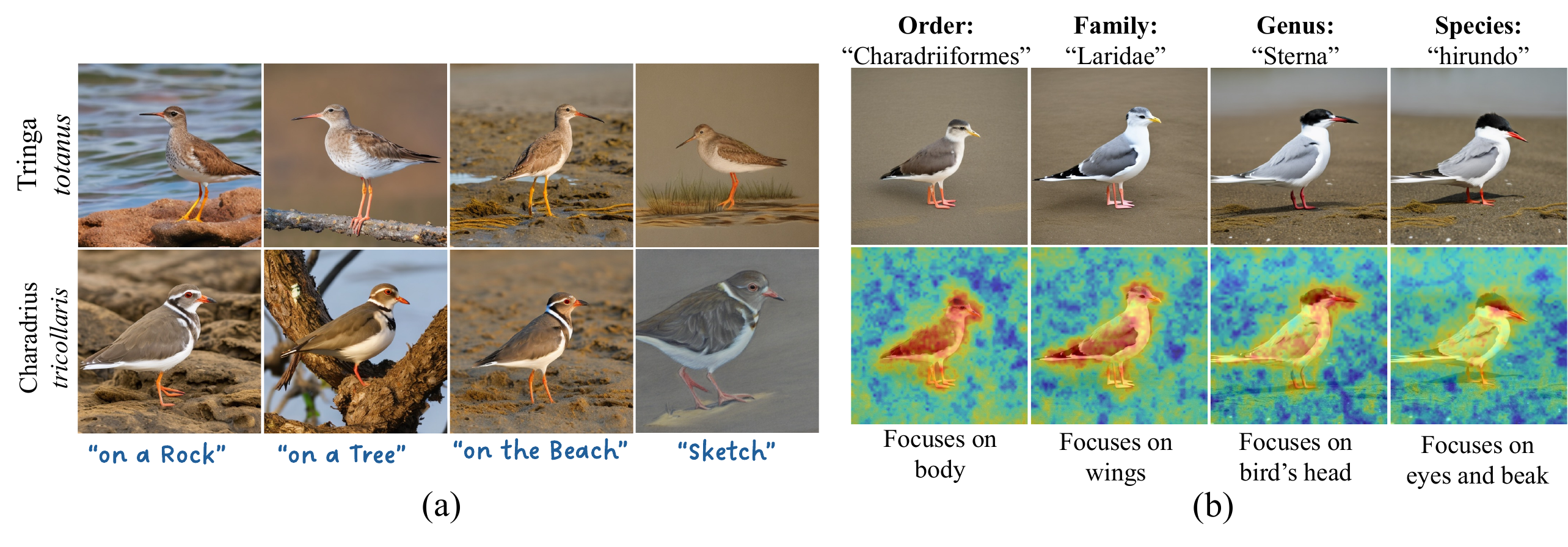}
\vspace{-5mm}
\caption{\small
(a) Free-form text prompting. Our dual-conditioning design allows flexible control over contextual attributes while preserving species-level morphology.
(b) Attention maps across different taxonomic tokens. \ourmethod gradually focuses on fine-grained information from body structure to head at \code{Genus} level to eyes and beak at \code{Species} level.
}
\label{fig:ablation_merged}
\vspace{-3mm}
\end{figure}

\mypara{Does Dual Conditioning Preserve Text Control?}
Biologists often rely on standardized illustrations, such as schematic sketches, that are commonly used to document and study trait variations~\cite{article, Wittmann_2013}.
Previous methods~\cite{monsefi2025taxadiffusion, pan2024finediffusion, khurana2024hierarchical} use taxonomy to fine-tune the diffusion models, which disrupts their ability to follow free-form text prompts. In contrast, \ourmethod retains the text pathway of the pretrained diffusion model. As shown in Figure~\ref{fig:ablation_merged}-(a) and Appendix~\ref{sup:free_form_text}, \ourmethod generates morphologically correct species across diverse contexts such as environment, pose, and styles.

\mypara{Can \ourmethod{} Enable Trait Discovery?}
We visualize cross-attention maps of \ourmethod{} for images generated at each taxonomic level~\cite{helbling2025conceptattention} in Figure~\ref{fig:ablation_merged}-(b). We observe a coarse-to-fine progression where coarse-taxonomic tokens (e.g., \code{Order}) emphasize global body structure, (\code{Family}) emphasizes on wing pattern, mid-level tokens (\code{Genus}) focus more on the head region, and \code{Species}-level tokens attend to fine-grained traits such as eyes and beak, generating novel scientific hypotheses about taxonomy-aligned traits that can be subsequently validated in future biological studies.


\begin{table}[t]
\centering

\begin{minipage}[t]{0.48\textwidth}
\centering
\small
\captionof{table}{\small Ablation study on Dual Conditions (DC) and Decoupled Cross-Attention (DCA) on \treeoflife-1M.}
\vspace{+1mm}

\resizebox{\linewidth}{!}{
\begin{tabular}{cccccccc}
\toprule
& DC & DCA & {FID$\downarrow$} & {LPIPS$\downarrow$} & {CAS@1$\uparrow$} & {CAS@5$\uparrow$} \\
\midrule
(a) & \xmark & \cmark & 35.52 & 0.735 &  56.42\% & 78.68\%  \\
(b) & \cmark & \xmark & 35.41 & 0.740 &  48.36\% & 72.86\%  \\
(c) & \cmark & \cmark & \textbf{29.87} & \textbf{0.734} &  \textbf{61.98\%} & \textbf{82.96\%} \\
\bottomrule
\end{tabular}
}
\label{tab:ablation}
\end{minipage}
\hfill
\begin{minipage}[t]{0.48\textwidth}
\centering
\small
\captionof{table}{\small Analysis of LoRA fine-tuning of SD 1.5, replacing CLIP with BioCLIP. Models are trained on iNat-mini-birds and evaluated on 650 bird species.}
\resizebox{\linewidth}{!}{
    \begin{tabular}{llccc}
      \toprule
      & Model & {FID} $\downarrow$ & {LPIPS} $\downarrow$ \\
      \midrule
      (a) & SD1.5 (ZS)   &  55.38 & 0.769 \\
      (b) & SD1.5 (CLIP) + LoRA  &  24.74 & 0.745  \\
      (c) & SD1.5 (BioCLIP) + LoRA  & 74.60  & 0.745  \\
      \midrule
      (d) & TaxaAdapter  & 17.08  & 0.745  \\
      \bottomrule
    \end{tabular}
    }
\label{tab:bioclip_replace_clip}
\end{minipage}
\vspace{-3mm}
\end{table}

\subsection{Ablation Analyses}
\label{sec:ablations}

\mypara{Benefits of Dual Conditions.}
We examine the effects of removing the text branch and conditioning only on VTM's taxonomy features. As shown in Table~\ref{tab:ablation} row (a), this single-stream variant underperforms the dual-conditioning design of row (c), suggesting complementary contextual information from the text pathway that boosts generation performance. 

\mypara{Effect of Decoupled Cross-Attention.}
We evaluate the role of decoupled cross-attention by replacing it with a single cross-attention layer that averages the text and taxonomy embeddings. Table~\ref{tab:ablation} row (b) shows a clear performance drop, suggesting that a unified attention stream cannot disentangle morphology from contextual cues. 

\mypara{Replacing CLIP with BioCLIP.}
As highlighted in Section~\ref{sec:non-trivial}, we show that na\"ive fine-tuning with Taxonomy (Table~\ref{tab:bioclip_replace_clip}-(b)) and replacing SD 1.5 CLIP's text encoder with a BioCLIP (Table~\ref{tab:bioclip_replace_clip}-(c)) is non-trivial, as diffusion backbones are heavily pretrained for text inputs and not VTM embeddings, and this interface mismatch that leads to inferior generation quality. See Appendix~\ref{sup:vtm_variants}.

\mypara{Why do we need Domain-Specific Encoders?}
We replace the VTM encoder in \ourmethod{} with a generic CLIP text encoder (while retaining the dual-conditioning design) and observe a performance drop in Table~\ref{tab:vtm_compatibility} row (a). This highlights that taxonomy–image aligned embeddings are critical for fine-grained species generation, and fine-tuning alone is not sufficient. 
Table~\ref{tab:vtm_compatibility} also shows compatibility with different VTMs.

\section{Conclusion and Future Work}
\label{sec:disc}
We introduce \ourmethod, a lightweight plug-in that injects VTM-derived taxonomy embeddings into frozen diffusion backbones, enabling scalable fine-grained species synthesis while preserving morphology fidelity, species identity, and full controllability through free-form text prompts. Future directions include extending \ourmethod to diffusion-transformer architectures and leveraging our MLLM-driven caption framework to interpret the semantic traits and distinctions across species and higher taxonomic groups. \ourmethod offers a scalable path toward taxonomy-aware generative modeling over the Tree of Life that continually improves as generative models evolve. We discuss the \textbf{Limitations and Failure Cases} of \ourmethod in Appendix~\ref{sup:limitations}.

\begin{ack}
Our research is supported in part by NSF OAC 2118240, NSF Award 2330423, NSERC Award 585136, the MIT Generative AI Consortium, and the NVIDIA Academic Grant Program. The authors acknowledge the Ohio Supercomputer Center and Advanced Research Computing at Virginia Tech for providing computational resources.
\end{ack}

{
    \small
    \bibliographystyle{plainnat}
    \bibliography{main}

@String(ICCV  = {Int. Conf. Comput. Vis.})

@String(ICLR  = {Int. Conf. Learn. Represent.})

@String(AAAI  = {AAAI})

@String(ICCV  = {ICCV})

@String(ICLR  = {ICLR})

@article{labs2025flux,
  title={FLUX. 1 Kontext: Flow Matching for In-Context Image Generation and Editing in Latent Space},
  author={Labs, Black Forest and Batifol, Stephen and Blattmann, Andreas and Boesel, Frederic and Consul, Saksham and Diagne, Cyril and Dockhorn, Tim and English, Jack and English, Zion and Esser, Patrick and others},
  journal={arXiv preprint arXiv:2506.15742},
  year={2025}
}

@article{hu2022lora,
  title={Lora: Low-rank adaptation of large language models.},
  author={Hu, Edward J and Shen, Yelong and Wallis, Phillip and Allen-Zhu, Zeyuan and Li, Yuanzhi and Wang, Shean and Wang, Liang and Chen, Weizhu and others},
  journal={Iclr},
  volume={1},
  number={2},
  pages={3},
  year={2022}
}

@article{kynkaanniemi2019improved,
  title={Improved precision and recall metric for assessing generative models},
  author={Kynk{\"a}{\"a}nniemi, Tuomas and Karras, Tero and Laine, Samuli and Lehtinen, Jaakko and Aila, Timo},
  journal={Advances in neural information processing systems},
  volume={32},
  year={2019}
}

@inproceedings{zhang2025biocap,
  title={BioCAP: exploiting synthetic captions beyond labels in biological foundation models},
  author={Zhang, Ziheng and Ma, Xinyue and Chowdhury, Arpita and Campolongo, Elizabeth and Thompson, Matthew and Zhang, Net and Stevens, Samuel and Lapp, Hilmar and Berger-Wolf, Tanya and Su, Yu and others},
  booktitle={International Conference on Learning Representations},
  volume={2026},
  pages={125415--125451},
  year={2026}
}

@misc{xlabs_flux_ip_adapter,
  title        = {FLUX IP-Adapter},
  author       = {{XLabs AI}},
  year         = {2024},
  howpublished = {\url{https://huggingface.co/XLabs-AI/flux-ip-adapter}},
  note         = {Hugging Face model repository}
}

@inproceedings{crispr,
title =  {TALEN and CRISPR/Cas Genome Editing Systems: Tools of Discovery.},
author ={Nemudryi, A. A. and Valetdinova, K. R. and Medvedev, S. P. and Zakian, S. M. },
booktitle={Acta naturae, 6(3)},
pages={19–40},
year={2014}
}

@article{raffel2020exploring,
  title={Exploring the limits of transfer learning with a unified text-to-text transformer},
  author={Raffel, Colin and Shazeer, Noam and Roberts, Adam and Lee, Katherine and Narang, Sharan and Matena, Michael and Zhou, Yanqi and Li, Wei and Liu, Peter J},
  journal={Journal of machine learning research},
  volume={21},
  number={140},
  pages={1--67},
  year={2020}
}

@article{helbling2025conceptattention,
  title={Conceptattention: Diffusion transformers learn highly interpretable features},
  author={Helbling, Alec and Meral, Tuna Han Salih and Hoover, Ben and Yanardag, Pinar and Chau, Duen Horng},
  journal={arXiv preprint arXiv:2502.04320},
  year={2025}
}

@article{Wittmann_2013, 
    title={Outlining Species: Drawing as a Research Technique in Contemporary Biology}, 
    volume={26}, DOI={10.1017/S0269889713000094}, 
    number={2}, 
    journal={Science in Context}, 
    author={Wittmann, Barbara}, 
    year={2013}, 
    pages={363–391}
}

@inproceedings{esser2024scaling,
  title={Scaling rectified flow transformers for high-resolution image synthesis},
  author={Esser, Patrick and Kulal, Sumith and Blattmann, Andreas and Entezari, Rahim and M{\"u}ller, Jonas and Saini, Harry and Levi, Yam and Lorenz, Dominik and Sauer, Axel and Boesel, Frederic and others},
  booktitle={Forty-first international conference on machine learning},
  year={2024}
}

@inproceedings{zhang2023adding,
  title={Adding conditional control to text-to-image diffusion models},
  author={Zhang, Lvmin and Rao, Anyi and Agrawala, Maneesh},
  booktitle={Proceedings of the IEEE/CVF international conference on computer vision},
  pages={3836--3847},
  year={2023}
}

@article{yang2024biotrove,
  title={Biotrove: A large curated image dataset enabling ai for biodiversity},
  author={Yang, Chih-Hsuan and Feuer, Benjamin and Jubery, Talukder and Deng, Zi and Nakkab, Andre and Hasan, Md Zahid and Chiranjeevi, Shivani and Marshall, Kelly and Baishnab, Nirmal and Singh, Asheesh and others},
  journal={Advances in Neural Information Processing Systems},
  volume={37},
  pages={102101--102120},
  year={2024}
}

@article{ravuri2019classification,
  title={Classification accuracy score for conditional generative models},
  author={Ravuri, Suman and Vinyals, Oriol},
  journal={Advances in neural information processing systems},
  volume={32},
  year={2019}
}

@misc{chen2023pixartalpha,
    title={PixArt-$\alpha$: Fast Training of Diffusion Transformer for Photorealistic Text-to-Image Synthesis}, 
    author={Junsong Chen and Jincheng Yu and Chongjian Ge and Lewei Yao and Enze Xie and Yue Wu and Zhongdao Wang and James Kwok and Ping Luo and Huchuan Lu and Zhenguo Li},
    year={2023},
    eprint={2310.00426},
    archivePrefix={arXiv},
    primaryClass={cs.CV}
}

@article{gu2025bioclip,
  title={Bioclip 2: Emergent properties from scaling hierarchical contrastive learning},
  author={Gu, Jianyang and Stevens, Samuel and Campolongo, Elizabeth G and Thompson, Matthew J and Zhang, Net and Wu, Jiaman and Kopanev, Andrei and Mai, Zheda and White, Alexander E and Balhoff, James and others},
  journal={arXiv preprint arXiv:2505.23883},
  year={2025}
}

@inproceedings{khan2023fishnet,
  title={Fishnet: A large-scale dataset and benchmark for fish recognition, detection, and functional trait prediction},
  author={Khan, Faizan Farooq and Li, Xiang and Temple, Andrew J and Elhoseiny, Mohamed},
  booktitle={Proceedings of the IEEE/CVF international conference on computer vision},
  pages={20496--20506},
  year={2023}
}

@inproceedings{adamkiewicz2026pretty,
  title={When Pretty Isn't Useful: Investigating Why Modern Text-to-Image Models Fail as Reliable Training Data Generators},
  author={Adamkiewicz, Krzysztof and Moser, Brian B and Frolov, Stanislav and Nauen, Tobias Christian and Raue, Federico and Dengel, Andreas},
  booktitle={Proceedings of the IEEE/CVF Conference on Computer Vision and Pattern Recognition},
  pages={36660--36669},
  year={2026}
}

@inproceedings{van2018inaturalist,
  title={The inaturalist species classification and detection dataset},
  author={Van Horn, Grant and Mac Aodha, Oisin and Song, Yang and Cui, Yin and Sun, Chen and Shepard, Alex and Adam, Hartwig and Perona, Pietro and Belongie, Serge},
  booktitle={Proceedings of the IEEE conference on computer vision and pattern recognition},
  pages={8769--8778},
  year={2018}
}

@article{van2017devil,
  title={The devil is in the tails: Fine-grained classification in the wild},
  author={Van Horn, Grant and Perona, Pietro},
  journal={arXiv preprint arXiv:1709.01450},
  year={2017}
}

@techreport{WahCUB_200_2011,
	Title = {Caltech-UCSD Birds-200-2011 (CUB-200-2011)},
	Author = {Wah, C. and Branson, S. and Welinder, P. and Perona, P. and Belongie, S.},
	Year = {2011},
	Institution = {California Institute of Technology},
	Number = {CNS-TR-2011-001}
}

@inproceedings{hessel2021clipscore,
  title={Clipscore: A reference-free evaluation metric for image captioning},
  author={Hessel, Jack and Holtzman, Ari and Forbes, Maxwell and Le Bras, Ronan and Choi, Yejin},
  booktitle={Proceedings of the 2021 conference on empirical methods in natural language processing},
  pages={7514--7528},
  year={2021}
}

@article{grattafiori2024llama,
  title={The llama 3 herd of models},
  author={Grattafiori, Aaron and Dubey, Abhimanyu and Jauhri, Abhinav and Pandey, Abhinav and Kadian, Abhishek and Al-Dahle, Ahmad and Letman, Aiesha and Mathur, Akhil and Schelten, Alan and Vaughan, Alex and others},
  journal={arXiv preprint arXiv:2407.21783},
  year={2024}
}

@article{zhu2025internvl3,
  title={Internvl3: Exploring advanced training and test-time recipes for open-source multimodal models},
  author={Zhu, Jinguo and Wang, Weiyun and Chen, Zhe and Liu, Zhaoyang and Ye, Shenglong and Gu, Lixin and Tian, Hao and Duan, Yuchen and Su, Weijie and Shao, Jie and others},
  journal={arXiv preprint arXiv:2504.10479},
  year={2025}
}

@misc{jiang2023mistral7b,
      title={Mistral 7B}, 
      author={Albert Q. Jiang and Alexandre Sablayrolles and Arthur Mensch and Chris Bamford and Devendra Singh Chaplot and Diego de las Casas and Florian Bressand and Gianna Lengyel and Guillaume Lample and Lucile Saulnier and Lélio Renard Lavaud and Marie-Anne Lachaux and Pierre Stock and Teven Le Scao and Thibaut Lavril and Thomas Wang and Timothée Lacroix and William El Sayed},
      year={2023},
      eprint={2310.06825},
      archivePrefix={arXiv},
      primaryClass={cs.CL},
      url={https://arxiv.org/abs/2310.06825}, 
}

@inproceedings{lin-2004-rouge,
    title = "{ROUGE}: A Package for Automatic Evaluation of Summaries",
    author = "Lin, Chin-Yew",
    booktitle = "Text Summarization Branches Out",
    month = jul,
    year = "2004",
    address = "Barcelona, Spain",
    publisher = "Association for Computational Linguistics",
    url = "https://aclanthology.org/W04-1013/",
    pages = "74--81"
}

@article{zhang2019bertscore,
  title={Bertscore: Evaluating text generation with bert},
  author={Zhang, Tianyi and Kishore, Varsha and Wu, Felix and Weinberger, Kilian Q and Artzi, Yoav},
  journal={arXiv preprint arXiv:1904.09675},
  year={2019}
}

@misc{flux2024,
    author={Black Forest Labs},
    title={FLUX},
    year={2024},
    howpublished={\url{https://github.com/black-forest-labs/flux}},
}

@article{podell2023sdxl,
  title={Sdxl: Improving latent diffusion models for high-resolution image synthesis},
  author={Podell, Dustin and English, Zion and Lacey, Kyle and Blattmann, Andreas and Dockhorn, Tim and M{\"u}ller, Jonas and Penna, Joe and Rombach, Robin},
  journal={arXiv preprint arXiv:2307.01952},
  year={2023}
}

@inproceedings{mou2024t2i,
  title={T2i-adapter: Learning adapters to dig out more controllable ability for text-to-image diffusion models},
  author={Mou, Chong and Wang, Xintao and Xie, Liangbin and Wu, Yanze and Zhang, Jian and Qi, Zhongang and Shan, Ying},
  booktitle={Proceedings of the AAAI conference on artificial intelligence},
  volume={38},
  number={5},
  pages={4296--4304},
  year={2024}
}

@inproceedings{stevens2024bioclip,
  title={Bioclip: A vision foundation model for the tree of life},
  author={Stevens, Samuel and Wu, Jiaman and Thompson, Matthew J and Campolongo, Elizabeth G and Song, Chan Hee and Carlyn, David Edward and Dong, Li and Dahdul, Wasila M and Stewart, Charles and Berger-Wolf, Tanya and others},
  booktitle={Proceedings of the IEEE/CVF conference on computer vision and pattern recognition},
  pages={19412--19424},
  year={2024}
}

@inproceedings{radford2021learning,
  title={Learning transferable visual models from natural language supervision},
  author={Radford, Alec and Kim, Jong Wook and Hallacy, Chris and Ramesh, Aditya and Goh, Gabriel and Agarwal, Sandhini and Sastry, Girish and Askell, Amanda and Mishkin, Pamela and Clark, Jack and others},
  booktitle={International conference on machine learning},
  pages={8748--8763},
  year={2021},
  organization={PmLR}
}

@article{ye2023ip,
  title={Ip-adapter: Text compatible image prompt adapter for text-to-image diffusion models},
  author={Ye, Hu and Zhang, Jun and Liu, Sibo and Han, Xiao and Yang, Wei},
  journal={arXiv preprint arXiv:2308.06721},
  year={2023}
}

@article{navard2024knobgen,
  title={KnobGen: controlling the sophistication of artwork in sketch-based diffusion models},
  author={Navard, Pouyan and Monsefi, Amin Karimi and Zhou, Mengxi and Chao, Wei-Lun and Yilmaz, Alper and Ramnath, Rajiv},
  journal={arXiv preprint arXiv:2410.01595},
  year={2024}
}

@inproceedings{rombach2022high,
  title={High-resolution image synthesis with latent diffusion models},
  author={Rombach, Robin and Blattmann, Andreas and Lorenz, Dominik and Esser, Patrick and Ommer, Bj{\"o}rn},
  booktitle={Proceedings of the IEEE/CVF conference on computer vision and pattern recognition},
  pages={10684--10695},
  year={2022}
}

@inproceedings{monsefi2025taxadiffusion,
  title={Taxadiffusion: Progressively trained diffusion model for fine-grained species generation},
  author={Monsefi, Amin Karimi and Khurana, Mridul and Ramnath, Rajiv and Karpatne, Anuj and Chao, Wei-Lun and Zhang, Cheng},
  booktitle={2025 IEEE/CVF International Conference on Computer Vision (ICCV)},
  pages={8579--8589},
  year={2025},
  organization={IEEE}
}

@inproceedings{zhang2023iti,
  title={Iti-gen: Inclusive text-to-image generation},
  author={Zhang, Cheng and Chen, Xuanbai and Chai, Siqi and Wu, Chen Henry and Lagun, Dmitry and Beeler, Thabo and De la Torre, Fernando},
  booktitle={Proceedings of the IEEE/CVF International Conference on Computer Vision},
  pages={3969--3980},
  year={2023}
}

@inproceedings{khurana2024hierarchical,
  title={Hierarchical Conditioning of Diffusion Models Using Tree-of-Life for Studying Species Evolution},
  author={Khurana, Mridul and Daw, Arka and Maruf, M and Uyeda, Josef C and Dahdul, Wasila and Charpentier, Caleb and Bak{\i}{\c{s}}, Yasin and Bart Jr, Henry L and Mabee, Paula M and Lapp, Hilmar and others},
  booktitle={European Conference on Computer Vision},
  pages={137--153},
  year={2024},
  organization={Springer}
}

@article{ho2022classifier,
  title={Classifier-free diffusion guidance},
  author={Ho, Jonathan and Salimans, Tim},
  journal={arXiv preprint arXiv:2207.12598},
  year={2022}
}

@article{pan2024finediffusion,
  title={FineDiffusion: Scaling up Diffusion Models for Fine-grained Image Generation with 10,000 Classes},
  author={Pan, Ziying and Wang, Kun and Li, Gang and He, Feihong and Lai, Yongxuan},
  journal={arXiv preprint arXiv:2402.18331},
  year={2024}
}

@inproceedings{li2025controlnet,
  title={ControlNet++: Improving Conditional Controls with Efficient Consistency Feedback},
  author={Li, Ming and Yang, Taojiannan and Kuang, Huafeng and Wu, Jie and Wang, Zhaoning and Xiao, Xuefeng and Chen, Chen},
  booktitle={European Conference on Computer Vision},
  pages={129--147},
  year={2025},
  organization={Springer}
}

@article{matsunaga2022fine,
  title={Fine-grained image editing by pixel-wise guidance using diffusion models},
  author={Matsunaga, Naoki and Ishii, Masato and Hayakawa, Akio and Suzuki, Kenji and Narihira, Takuya},
  journal={arXiv preprint arXiv:2212.02024},
  year={2022}
}

@article{huang2024diffusion,
  title={Diffusion model-based image editing: A survey},
  author={Huang, Yi and Huang, Jiancheng and Liu, Yifan and Yan, Mingfu and Lv, Jiaxi and Liu, Jianzhuang and Xiong, Wei and Zhang, He and Chen, Shifeng and Cao, Liangliang},
  journal={arXiv preprint arXiv:2402.17525},
  year={2024}
}

@inproceedings{kim2022diffusionclip,
  title={Diffusionclip: Text-guided diffusion models for robust image manipulation},
  author={Kim, Gwanghyun and Kwon, Taesung and Ye, Jong Chul},
  booktitle={Proceedings of the IEEE/CVF conference on computer vision and pattern recognition},
  pages={2426--2435},
  year={2022}
}

@article{yue2023chatface,
  title={Chatface: Chat-guided real face editing via diffusion latent space manipulation},
  author={Yue, Dongxu and Guo, Qin and Ning, Munan and Cui, Jiaxi and Zhu, Yuesheng and Yuan, Li},
  journal={arXiv preprint arXiv:2305.14742},
  year={2023}
}

@article{han2023highly,
  title={Highly personalized text embedding for image manipulation by stable diffusion},
  author={Han, Inhwa and Yang, Serin and Kwon, Taesung and Ye, Jong Chul},
  journal={arXiv preprint arXiv:2303.08767},
  year={2023}
}

@article{zhang2024survey,
  title={A Survey on Personalized Content Synthesis with Diffusion Models},
  author={Zhang, Xulu and Wei, Xiao-Yong and Zhang, Wengyu and Wu, Jinlin and Zhang, Zhaoxiang and Lei, Zhen and Li, Qing},
  journal={arXiv preprint arXiv:2405.05538},
  year={2024}
}

@inproceedings{ruiz2023dreambooth,
  title={Dreambooth: Fine tuning text-to-image diffusion models for subject-driven generation},
  author={Ruiz, Nataniel and Li, Yuanzhen and Jampani, Varun and Pritch, Yael and Rubinstein, Michael and Aberman, Kfir},
  booktitle={Proceedings of the IEEE/CVF conference on computer vision and pattern recognition},
  pages={22500--22510},
  year={2023}
}

@inproceedings{li2024stylegan,
  title={When stylegan meets stable diffusion: a w+ adapter for personalized image generation},
  author={Li, Xiaoming and Hou, Xinyu and Loy, Chen Change},
  booktitle={Proceedings of the IEEE/CVF Conference on Computer Vision and Pattern Recognition},
  pages={2187--2196},
  year={2024}
}

@article{wang2024animatelcm,
  title={Animatelcm: Accelerating the animation of personalized diffusion models and adapters with decoupled consistency learning},
  author={Wang, Fu-Yun and Huang, Zhaoyang and Shi, Xiaoyu and Bian, Weikang and Song, Guanglu and Liu, Yu and Li, Hongsheng},
  journal={arXiv preprint arXiv:2402.00769},
  year={2024}
}

@inproceedings{ruiz2024hyperdreambooth,
  title={Hyperdreambooth: Hypernetworks for fast personalization of text-to-image models},
  author={Ruiz, Nataniel and Li, Yuanzhen and Jampani, Varun and Wei, Wei and Hou, Tingbo and Pritch, Yael and Wadhwa, Neal and Rubinstein, Michael and Aberman, Kfir},
  booktitle={Proceedings of the IEEE/CVF Conference on Computer Vision and Pattern Recognition},
  pages={6527--6536},
  year={2024}
}

@misc{Seitzer2020FID,
  author={Maximilian Seitzer},
  title={{pytorch-fid: FID Score for PyTorch}},
  month={August},
  year={2020},
  note={Version 0.3.0},
  howpublished={\url{https://github.com/mseitzer/pytorch-fid}},
}

@inproceedings{sastry2025taxabind,
  title={Taxabind: A unified embedding space for ecological applications},
  author={Sastry, Srikumar and Khanal, Subash and Dhakal, Aayush and Ahmad, Adeel and Jacobs, Nathan},
  booktitle={2025 IEEE/CVF Winter Conference on Applications of Computer Vision (WACV)},
  pages={1765--1774},
  year={2025},
  organization={IEEE}
}

@article{zhao2023uni,
  title={Uni-controlnet: All-in-one control to text-to-image diffusion models},
  author={Zhao, Shihao and Chen, Dongdong and Chen, Yen-Chun and Bao, Jianmin and Hao, Shaozhe and Yuan, Lu and Wong, Kwan-Yee K},
  journal={Advances in Neural Information Processing Systems},
  volume={36},
  pages={11127--11150},
  year={2023}
}

@inproceedings{li2024photomaker,
  title={Photomaker: Customizing realistic human photos via stacked id embedding},
  author={Li, Zhen and Cao, Mingdeng and Wang, Xintao and Qi, Zhongang and Cheng, Ming-Ming and Shan, Ying},
  booktitle={Proceedings of the IEEE/CVF conference on computer vision and pattern recognition},
  pages={8640--8650},
  year={2024}
}

@inproceedings{zhang2018unreasonable,
  title={The unreasonable effectiveness of deep features as a perceptual metric},
  author={Zhang, Richard and Isola, Phillip and Efros, Alexei A and Shechtman, Eli and Wang, Oliver},
  booktitle={Proceedings of the IEEE conference on computer vision and pattern recognition},
  pages={586--595},
  year={2018}
}

@article{article,
author = {Fisher, Ray and Dowling, Ashley},
year = {2010},
month = {10},
pages = {395-409},
title = {Modern methods and technology for doing classical taxonomy},
volume = {50},
journal = {Acarologia},
doi = {10.1051/acarologia/20101981}
}

@article{oquab2023dinov2,
  title={Dinov2: Learning robust visual features without supervision},
  author={Oquab, Maxime and Darcet, Timoth{\'e}e and Moutakanni, Th{\'e}o and Vo, Huy and Szafraniec, Marc and Khalidov, Vasil and Fernandez, Pierre and Haziza, Daniel and Massa, Francisco and El-Nouby, Alaaeldin and others},
  journal={arXiv preprint arXiv:2304.07193},
  year={2023}
}
}

\clearpage
\appendix

\appendix

\makeatletter
\let\origaddcontentsline\addcontentsline
\renewcommand{\addcontentsline}[3]{%
  \def\tempext{#1}%
  \def\tocext{toc}%
  \ifx\tempext\tocext
    \origaddcontentsline{atoc}{#2}{#3}%
  \else
    \origaddcontentsline{#1}{#2}{#3}%
  \fi
}
\makeatother

\section*{Appendix}

\begingroup
\setcounter{tocdepth}{2}
\hypersetup{
    colorlinks=true,
    linkcolor=black,
    citecolor=blue,
    urlcolor=blue,
    linktoc=all
}
\makeatletter
\@starttoc{atoc}
\makeatother
\endgroup

\gradientline

\section{Additional Implementation Details}
\label{sup:impl}

\subsection{Background and Additional Method Details}
\label{subsec:background}

\mypara{Text-to-Image Diffusion Models.}
Diffusion models generate samples by progressively denoising a noisy latent variable $\mathbf{z}_t$ at timestep $t$. In conditional generation, the denoising model is additionally conditioned on an auxiliary signal $\mathbf{c}$, such as a class label, text prompt, or image embedding, to guide the generation process~\cite{rombach2022high}. At each timestep, a denoising network predicts the noise added to the clean latent $\mathbf{z}_0$. The standard training objective is
\begin{equation}\label{eq:diffusion}
\mathcal{L}(\theta)
=
\mathbb{E}_{\mathbf{z}_0,\boldsymbol{\epsilon},\mathbf{c},t}
\left[
\left\lVert
\boldsymbol{\epsilon}
-
\boldsymbol{\epsilon}_\theta(\mathbf{z}_t,\mathbf{c},t)
\right\rVert_2^2
\right],
\end{equation}
where $\boldsymbol{\epsilon}$ is sampled Gaussian noise and $\boldsymbol{\epsilon}_\theta$ is the denoising network. In text-to-image diffusion models, the conditioning signal $\mathbf{c}$ is typically produced by a pretrained text encoder, such as CLIP~\cite{radford2021learning}, and injected into the denoising U-Net through cross-attention layers.

\mypara{Vision Taxonomy Models.}
Vision Taxonomy Models (VTMs) are trained to align visual observations with taxonomic information, producing embeddings that capture biological and taxonomic structure. Given a taxonomic hierarchy string
$y_{\text{taxa}}$ from \code{Kingdom} to \code{Species}, for example ``\emph{Animalia, Chordata, Aves, Passeriformes, Passerellidae, Spizella, breweri}'', a VTM taxonomy encoder $\mathcal{E}_{\text{VTM}}$ produces a taxonomy-aware embedding:
\begin{equation}\label{eq:ctaxa}
    \mathbf{c}_{\text{taxa}} = \mathcal{E}_{\text{VTM}}(y_{\text{taxa}}).
\end{equation}

\mypara{Taxonomy-Text Dual Conditioning.}
To combine species-level taxonomic information with the flexible controllability of natural-language prompts, we use a taxonomy-text dual-conditioning design. The model receives two complementary conditioning signals:
\begin{equation}\label{eq:ctaxa_ctext}
    \mathbf{c}_{\text{taxa}} = \mathcal{E}_{\text{VTM}}(y_{\text{taxa}}),
    \qquad
    \mathbf{c}_{\text{text}} = \mathcal{E}_{\text{CLIP}}(y_{\text{text}}),
\end{equation}

where $y_{\text{text}}$ is a natural-language prompt such as ``\emph{a photo of a bird}''. In practice, we incorporate taxonomy and adopt the template `\emph{a photo of} $y_{\text{taxa}}$'' to maintain semantic consistency with the introduced embeddings.

\mypara{Decoupled Cross-Attention.}
 We repurpose the decoupled cross-attention mechanism from IP-adapter~\cite{ye2023ip} that assigns a separate attention to each conditioning source. Specifically, the taxonomy embedding $\mathbf{c}_\text{taxa}$ is first passed through a projection network consisting of a linear layer and LayerNorm. Let $\bm{Q}$ be the query from the U-Net features, and $(\bm{K}_\text{text}, \bm{V}_\text{text})$ and $(\bm{K}_\text{taxa},\bm{V}_\text{taxa})$ be key-value pairs from the text and taxonomy branches, respectively. The dual-attention update is written as follows:
\begin{equation}\label{eq:dual_attention}
\bm{Z} = \mathrm{Att}(\bm{Q}, \bm{K}_\text{text}, \bm{V}_\text{text}) + \mathrm{Att}(\bm{Q}, \bm{K}_\text{taxa}, \bm{V}_\text{taxa})  
\end{equation}
Our use of decoupled cross-attention allows the taxonomy branch to focus on species-level morphology, while the text branch controls contextual cues such as style and background, enabling complementary conditioning.

\mypara{Parameter-Efficient Training.}
During training, we update only the projection network and the decoupled cross-attention layers, keeping all parameters of the pretrained diffusion model frozen. 
The model is trained on large-scale taxonomy–image pairs using a similar denoising objective as in Equation~\ref{eq:diffusion}:
\begin{equation}
\mathbb{E}_{\mathbf{z}_0,\boldsymbol{\epsilon},\mathbf{c}_\text{text}, \mathbf{c}_\text{taxa},t}\!\left[
\left\lVert \boldsymbol{\epsilon}-\boldsymbol{\epsilon}_\theta(\mathbf{z}_t,\mathbf{c}_\text{text},\mathbf{c}_\text{taxa}, t)\right\rVert_2^2
\right],
\end{equation}
To enable Classifier-Free Guidance~\cite{ho2022classifier} at inference, we randomly drop the dual-conditioning branch during training. The parameter-efficient strategy minimizes computational cost and prevents forgetting of the pretrained text–image alignment, allowing the model to retain its ability to be controlled by text prompts while incorporating biological knowledge.

\mypara{Inference.}
At inference time, both conditioning streams operate jointly to guide the denoising process.
Since the taxonomy and text cross-attention modules are decoupled, their contributions can be independently controlled by a weighting factor $\lambda$:
\begin{equation}\label{eq:inference}
\bm{Z} = \mathrm{Att}(\bm{Q}, \bm{K}_\text{text}, \bm{V}_\text{text}) + \lambda \cdot \mathrm{Att}(\bm{Q}, \bm{K}_\text{taxa}, \bm{V}_\text{taxa})
\end{equation}
where $\lambda$ modulates taxonomy conditioning strength.
Setting $\lambda=0$ recovers the original text-to-image diffusion model, while a larger $\lambda$ emphasizes taxonomy-specific traits derived from VTMs.
This provides intuitive control between visual appearance and biological precision for fine-grained species generation.

\mypara{Free-form Text Prompting.}
Thanks to the taxonomy-text dual conditioning design, we observe that the text branch remains fully functional and can be freely customized to specify background, pose, lighting, or artistic style.
For example, given a taxonomy sequence $y_\text{taxa}$, users can prompt ``$y_\text{taxa}$ \emph{on a tree branch}'', ``$y_\text{taxa}$ \emph{under cloudy sky}'' or ``$y_\text{taxa}$ \emph{as a schematic sketch}'' to produce morphologically accurate yet stylistically diverse outputs.
This flexibility makes \ourmethod suitable for scientific visualization, while the taxonomy branch ensures trait fidelity.

\subsection{Backbone and Conditioning Modules}
\label{sup:impl:backbone}

\mypara{Diffusion Backbone.}
All experiments were conducted using a publicly available Stable Diffusion v1.5~\cite{rombach2022high} latent diffusion model backbone, and all the weights are frozen throughout training.
Images are resized and center-cropped to a resolution of $512 \times 512$, and encoded into a latent tensor of shape $64 \times 64 \times 4$. We also evaluate using Stable Diffusion XL~\cite{podell2023sdxl} and present its results in Section~\ref{sup:sdxl}.

\mypara{Taxonomy Encoder.}
Unless specified, the taxonomy encoder $\mathcal{E}_{\text{VTM}}$ is instantiated from BioCLIP 2~\cite{gu2025bioclip}.
We also evaluate our framework using TaxaBind~\cite{sastry2025taxabind} and BioTrove-CLIP~\cite{yang2024biotrove} as alternative Vision Taxonomy Models (VTMs). In all cases, the VTMs' taxonomy encoders are kept frozen.
We feed the full 7-level hierarchical Linnaean taxonomy paths defined as \code{Kingdom}, \code{Phylum}, \code{Class}, \code{Order}, \code{Family}, \code{Genus} and \code{Species} as a string (e.g., ``\emph{Animalia, Chordata, Aves, Passeriformes, Passerellidae, Spizella, breweri}'') into the taxonomy encoder $\mathcal{E}_{\text{VTM}}$ of the VTM.

\subsection{Training and Hyperparameter Details}
\label{sup:impl:hyper}

\mypara{Model Parameters.}
Our parameter-efficient training strategy (Section~\ref{subsec:background}) only updates the following:
(i) the projection network applied to $\mathbf{c}_{\text{taxa}}$ (Eq. \ref{eq:ctaxa_ctext}), and
(ii) the additional weights introduced by the taxonomy-decoupled cross-attention.
All other parameters of the diffusion backbone and VTMs remain frozen.
In total, this adds approximately $22$M trainable parameters, which is a small fraction of the 860 ($\sim$ 2.6\%) parameters of the full Stable Diffusion model.
This keeps memory and compute overhead modest while enabling strong taxonomy-aware control.

\mypara{Training Setup.}
Table~\ref{tab:sup-hyperparameters} summarizes the main hyperparameters used for \ourmethod.
All models are trained for 100 epochs with a batch size of 64 images per GPU. We use 4 NVIDIA H100 GPUs with mixed-precision (\texttt{fp16}) for training the models. Optimization is performed with AdamW, using a learning rate of $1\times10^{-4}$ and weight decay of $0.01$ for all adapter parameters. For classifier-free guidance, we follow the standard $\epsilon$-prediction objective in Eq.~\ref{eq:diffusion} and randomly drop the conditioning during training with probability $0.05$ and replace both $\mathbf{c}_{\text{text}}$ and $\mathbf{c}_{\text{taxa}}$ with null tokens (i.e., an unconditional branch). All training images are resized to $512 \times 512$ and taxonomic labels are taken from the dataset metadata.

\mypara{Compute.}
\ourmethod takes $\sim$1 day to train on the FishNet dataset, whereas full fine-tuning of SD1.5 requires updating $\sim$860M parameters, which takes $\sim$6 days. For LoRA baselines, we train \flux and SD3.5-Large with LoRA rank 16 for 50k iterations with the same batch size of 64 per GPU.

\begin{table}[t]
    \centering
    \small
    \tabcolsep 8pt
    \caption{\small Training and inference hyperparameter for \ourmethod. These settings are used across all datasets unless otherwise noted.}
    \label{tab:sup-hyperparameters}
    \begin{tabular}{l c}
    \toprule
    \textbf{Parameter} & \textbf{Value} \\
    \midrule
    Base model & Stable Diffusion v1.5 (frozen) \\
    VTM encoder & BioCLIP / TaxaBind / BioTrove-CLIP (frozen) \\
    Image resolution & $512 \times 512$ \\
    Latent shape & $64 \times 64 \times 4$ \\
    Trainable parameters & $\approx 22$M (proj + attn) \\
    Epochs & 100 \\
    Batch size per GPU & 64 \\
    Optimizer & AdamW \\
    Learning rate & $1\times10^{-4}$ \\
    Weight decay & $0.01$ \\
    Precision & \texttt{fp16} \\
    Classifier-free drop prob. & $0.05$ (dual conditioning) \\
    \bottomrule
    \end{tabular}
\end{table}

\mypara{Inference Setup.}
During inference, we use the dual-conditioning formulation from Eq.~\ref{eq:inference}. We fix $\lambda = 1$ for all quantitative results. This setting provides a good balance between preserving species-level morphology (taxonomy branch). For free-form text prompting, we use $\lambda = 0.5$ to allow flexible control as justified in Section~\ref{sup:free_form_text}.
We use a standard deterministic sampler with 50 denoising steps for all models, including baselines and a classifier-free guidance scale of 3.5. We run all our experiments with a fixed seed of 42.

\subsection{Trait Fidelity Evaluation}
\label{sup:trait_fidelity_appendix}

\mypara{Per-image trait captioning.} 
For each species, we prompt an InternVL3-8B~\cite{zhu2025internvl3} to describe only visible traits. We explicitly exclude species names to avoid bias. In our experiments, we sample 10 images per species for the generated set, resulting in 10 corresponding trait captions. For real images, we caption all the available images of that species. We use the following prompt for per-image trait captions:

\begin{tcolorbox}[colback=gray!5,colframe=black!40,boxrule=0.4pt,
                  left=6pt,right=6pt,top=6pt,bottom=6pt]
\small
\texttt{<image> Describe this image in detail. Focus on the main subject, its appearance, colors, shape, pattern, texture, position, and any distinctive features. Ignore the background.}
\end{tcolorbox}

\mypara{Per-species Trait Summarization.}
For generating a per-species trait summarization, we employ LLaMA3.1-8B-Instruct~\cite{grattafiori2024llama} to consolidate the per-image captions into a single ``trait caption'' for each species. We use the prompt:
\begin{tcolorbox}[colback=gray!5,colframe=black!40,boxrule=0.4pt,
                  left=6pt,right=6pt,top=6pt,bottom=6pt]
\small
\texttt{<captions> Each line is a caption from one image of this species. Produce one combined caption that reflects the shared visible traits, including colors, patterns, shapes, markings, textures, and other distinctive parts. Ignore the background.}
\end{tcolorbox}

\section{Revisiting Main Results}
\label{sup:main_results}

\mypara{Results on iNat-mini}.
Table~\ref{tab:inat} of the main paper presents results on the iNat-mini dataset. \ourmethod achieves consistent improvements across all evaluation metrics.
For classification accuracy, we train a linear classifier using pre-trained DINOv2 features using a separate 500k image subset sampled from the remaining 2.2M iNat images to avoid data leakage; the base classifier achieves 72.26\% top-1 accuracy on its validation split.
For caption-based trait fidelity, we report BERTScore, CLIPScore, and Mistral-7B similarity for semantic alignment, and ROUGE-L for word-level overlap of species-defining traits between real and generated caption sets. \ourmethod achieves stronger trait alignment than all baselines. 

We observe that SD1.5 outperforms the larger baselines. FID measures distributional similarity between generated images and the biodiversity reference dataset, whereas CAS (Classification Accuracy Score) measures how accurately the generated images preserve the prompted species identity. Consequently, lower FID does not necessarily imply stronger species fidelity. Across our results, SD3.5-Large and FLUX.1-dev generally achieve higher CAS than SD1.5, indicating better species-identity preservation despite worse FID. For example, on \treeoflife-1M, FLUX achieves higher CAS@1 than SD1.5 (12.44\% vs. 9.32\%) but substantially worse FID (90.09 vs. 46.89).

This interpretation is further supported by generative precision and recall. FLUX achieves higher precision than SD1.5 (0.66 vs. 0.54), indicating stronger sample fidelity, but substantially lower recall (0.06 vs. 0.33), indicating limited coverage of the reference distribution. One possible explanation is a compositional mismatch: \treeoflife-1M contains organisms across highly diverse natural settings, whereas larger open-domain models often produce polished, portrait-style, foreground-centric images with limited variation in pose, scale, and environment (see Figure~\ref{fig:qual_tol}). K. Adamkiewicz et al. \cite{adamkiewicz2026pretty} also show a similar finding that most recent text-to-image models collapse to generating a narrow, aesthetic-centric distribution that undermines diversity and real data distribution coverage. This mismatch possibly explains the larger models’ poorer FID despite their stronger CAS.

Prompt familiarity further contributes to the results because generic models may not be explicitly trained to interpret complete Linnaean taxonomy strings. As shown in Appendix \ref{sup:common_vs_taxonomic_name} and Table~\ref{tab:common-v-taxanomy}, replacing these strings with common species names improves FLUX’s FID from 90.09 to 68.29 and CAS@1 from 12.44\% to 21.28\%, confirming that familiarity with the prompt formulation substantially affects performance. Similar observations are made in Appendix~\ref{sup:biocap_t2i} and Table~\ref{tab:biocap_t2i} where BioCAP captions are used for prompting. We also tried prompting all models to generate organisms in natural scenes. It produced similar results, suggesting that this behavior cannot be resolved through natural-scene prompting alone.

\mypara{Results on {\sc \textbf{TreeOfLife}}-1M}.
We further evaluate on the larger \treeoflife-1M dataset, which covers a broader taxonomy span and species diversity, making it a substantially harder benchmark than iNat-mini. As observed in Table~\ref{tab:tol-1m-1}, \ourmethod maintains strong performance and generalizes effectively at scale, achieving substantial gains in both image quality and taxonomy–image alignment. Notably, the results on \treeoflife-1M surpass those on iNat-mini, suggesting that more diverse training data enhances fine-grained species generation.

\mypara{Results on FishNet}.
Table~\ref{tab:fishnet} of the main paper reports results on the fine-grained FishNet dataset. \ourmethod improves taxonomy–image alignment while attaining the lowest FID. We note a slight increase in LPIPS compared with TaxaDiffusion, which we hypothesize reflects the model’s stronger emphasis on fine-scale trait fidelity rather than global texture smoothness. This trade-off diminishes as training scales to larger datasets, consistent with the trend observed in Table~\ref{tab:inat} of the main paper.

\section{Additional Results}
\label{sup:additional_results}

\subsection{Trait Fidelity Evaluation}

\mypara{Results on {\sc \textbf{TreeOfLife}}-1M.}
Table~\ref{tab:tol-1m-1} in the main paper reports trait-fidelity scores on \treeoflife-1M using the MLLM-based pipeline described in Section~\ref{subsec:vlm_metric} of the main paper. For each species, we first generate image-level captions from InternVL3-8B and then summarize them into a single {trait caption} with LLaMA3.1-8B-Instruct. This per-species summarization is important for capturing stable morphological signatures: distinctive colors, patterns, or body parts may be occluded or absent in any single image due to pose, orientation, or partial visibility. Aggregating over multiple captions yields a more robust description of species-level traits and reduces noise from background clutter.

To disentangle the effect of this summarization step, Table~\ref{tab:tol-1m-captions-image-to-image} evaluates an instance-wise variant of the same pipeline. Here, we directly compare captions of real and generated images without summarization, compute similarity scores for each image pair, and then average over images within each species. Despite being more sensitive to per-image variability, this image-to-image setting shows the same ordering as the summarized variant. \ourmethod\ consistently outperforms all baselines across BERT Score, CLIP Score, ROUGE-L, and Mistral-7B text similarity. We additionally experimented with a one-to-many matching scheme, where each generated image caption is compared against all real image captions of the same species; this produced similar relative performance.

\begin{table}[h]
    \centering
    \tabcolsep 4.5pt
    \small
    \caption{\small Quantitative results on instance-wise trait-fidelity for \treeoflife-1M. Results are evaluated using 500 species on \treeoflife-1M.
    }
        \begin{tabular}{lcccc}
        \toprule
        Model & {BERT$\uparrow$} & {CLIP$\uparrow$} & {ROUGE-L$\uparrow$} & {Mistral-7B$\uparrow$} \\
        \midrule
        SD1.5       & 0.30 & 0.61 & 0.13 & 0.87\\
        SD3.5-Large & 0.28 & 0.61 & 0.13 & 0.87 \\
        \flux       & 0.30 & 0.62 & 0.13 & 0.87 \\
        \midrule
        \textbf{\ourmethod} & \textbf{0.35} & \textbf{0.67} & \textbf{0.31} & \textbf{0.89} \\
        \bottomrule
        \end{tabular}
    \label{tab:tol-1m-captions-image-to-image}
\end{table}

\mypara{Results on iNat-mini.}
Table~\ref{tab:image-to-image-captions} reports the instance-wise variant trait similarity as seen above on the iNat-mini dataset. Again, we observe that \ourmethod\ achieves the highest scores on all four caption-based metrics, indicating that its improvements are not limited to a particular dataset or evaluation variant. Taken together, the summarized and instance-wise results across \treeoflife-1M and iNat-mini provide consistent evidence that taxonomy-aware conditioning leads to captions that more faithfully capture species-defining visual traits than those produced by strong off-the-shelf text-to-image models.

\begin{table}[h]
    \centering
    \tabcolsep 4.5pt
    \small
    \caption{\small Quantitative results on instance-wise trait-fidelity for iNat-mini.
    }
        \begin{tabular}{lcccc}
        \toprule
        Model & {BERT$\uparrow$} & {CLIP$\uparrow$} & {ROUGE-L$\uparrow$} & {Mistral-7B$\uparrow$} \\
        \midrule
        SD1.5       & 0.32 & 0.66 & 0.29 & 0.89\\
        SD3.5-Large & 0.32 & 0.66 & 0.29 & 0.89 \\
        \flux       & 0.34 & 0.69 & 0.30 & 0.90 \\
        \midrule
        \textbf{\ourmethod} & \textbf{0.38} & \textbf{0.73} & \textbf{0.33} & \textbf{0.92} \\
        \bottomrule
        \end{tabular}
    \label{tab:image-to-image-captions}
\end{table}

\begin{figure}[h]
    \centering
    \includegraphics[width=\linewidth]{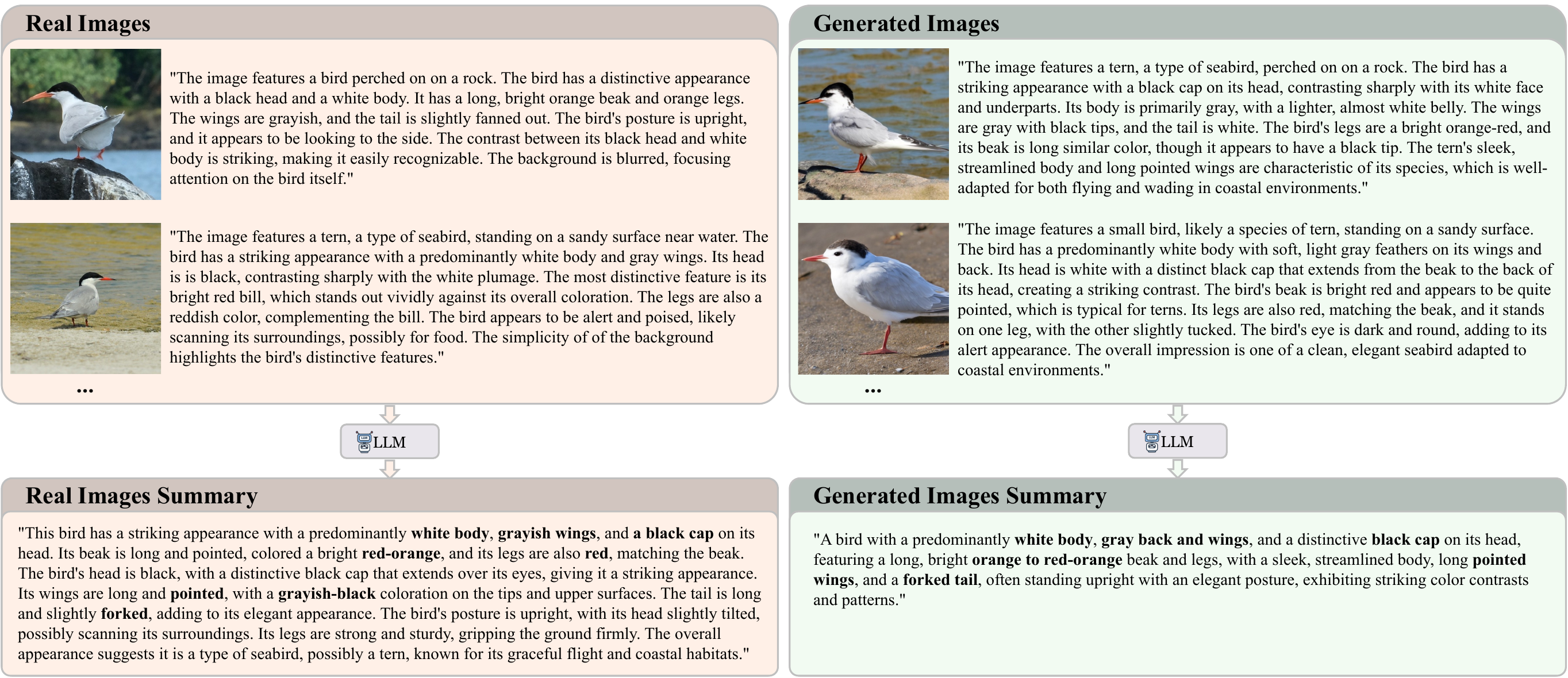}
    \vspace{1mm}
    \gradientseparator
    \vspace{1mm}
    \includegraphics[width=\linewidth]{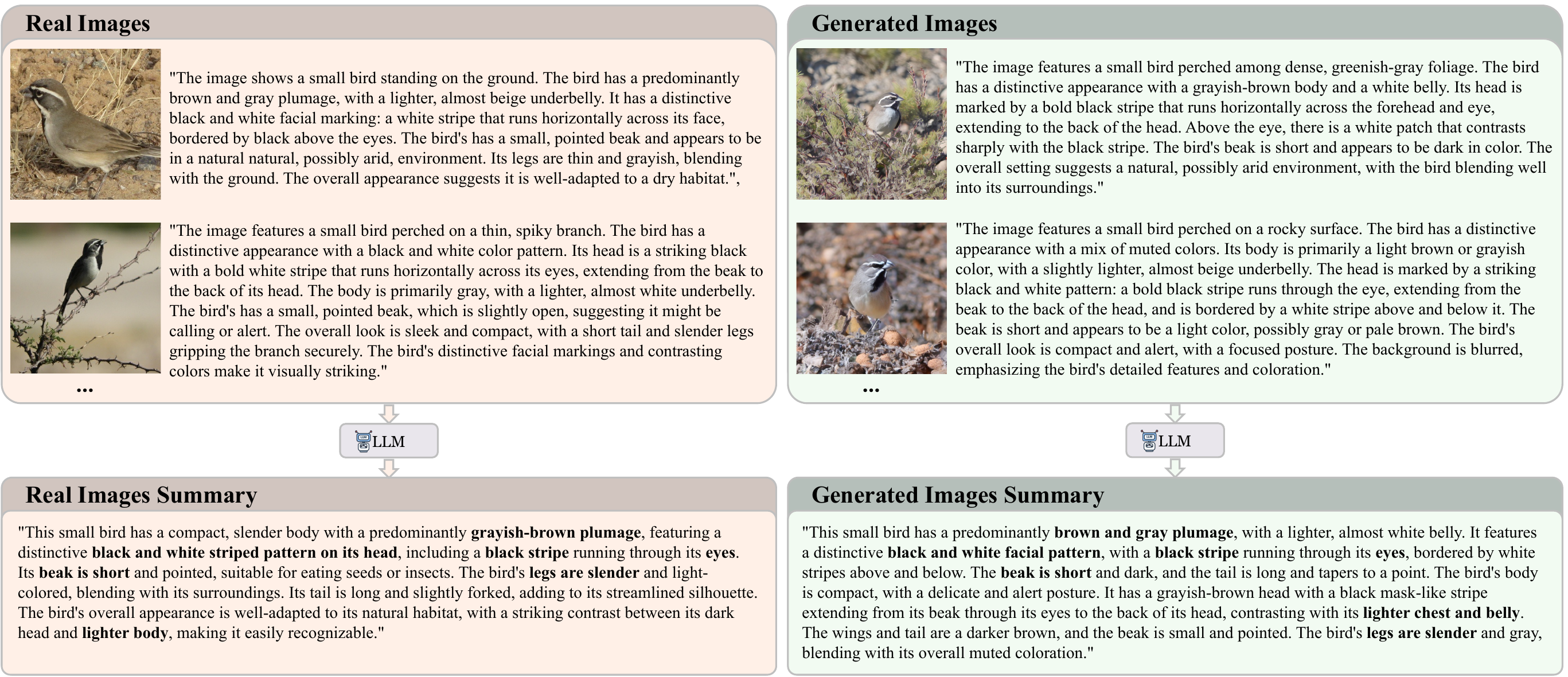}
    \caption{\small Qualitative results for trait fidelity analysis using our captioning-based evaluation method. The upper case shows the trait captions generated using MLLM for real and generated images for species ``\textit{Sterna hirundo}'' along with the summarized captions. The bottom case shows the trait-fidelity captions for species ``\textit{Amphispiza bilineata}''.}
    \label{fig:vlm-captions}
\end{figure}

\mypara{Examples of Trait Fidelity Evaluation.}
Figure~\ref{fig:vlm-captions} shows trait captioning results for species ``\textit{Animalia, Chordata, Aves, Charadriiformes, Laridae, Sterna hirundo}'' with the common name `Common tern' and ``\textit{Animalia, Chordata, Aves, Passeriformes, Passerellidae, Amphispiza, bilineata}'' with the common name `Black-throated Sparrow', respectively. We show the captions generated for different images and the trait-summarized captions across all the images for each species.

\subsection{User Study for Trait Fidelity Evaluation}
\label{sup:user_study}
To validate whether our generated text summaries are faithful to visual traits, we conduct a small human evaluation guided by curated Wikipedia descriptions. We sampled 10 species randomly from the iNat-mini dataset and, for each species, the evaluators were shown (i) a small set of real images for reference, and (ii) a ground-truth textual description extracted from Wikipedia that emphasizes visual traits (e.g., coloration patterns, shapes, \etc). We then show the MLLM-generated trait summary for that species and ask evaluators to rate its correctness and completeness with respect to the reference images and Wikipedia text. We had 12 evaluators, including evolutionary biologists and experts actively working at the intersection of machine learning and ecology/biology. The study yielded an average rating of $4.25/5.0$, suggesting that the generated summaries capture salient, visually grounded traits.

\subsection{Trait Fidelity Evaluation with BioCAP Captions as Biologically Informed External References}
\label{sup:biocap_grounding}
To further validate whether our caption-based trait-fidelity evaluation captures biologically meaningful visual traits, we compare generated-image trait summaries against BioCAP~\cite{zhang2025biocap} reference captions. Unlike our main evaluation, where real-image references are produced using the same MLLM-based pipeline, BioCAP provides independent trait-rich descriptive captions grounded in Wikipedia context and taxon-specific examples. This allows us to evaluate whether generated images align with external biologically informed trait descriptions rather than only with MLLM-generated summaries from real images. As shown in Table~\ref{tab:inat-caption-trait-fidelity}, \ourmethod achieves the best performance across all four metrics, suggesting that \ourmethod produces images whose inferred visual traits better align with biologically informed external references.

This consistency also supports our main evaluation protocol: using real images as reference trait ground truth through the same captioning pipeline yields rankings consistent with independent BioCAP references, making the proposed metric scalable to taxa for which curated trait descriptions are unavailable.

\begin{table}[h]
    \centering
    \tabcolsep 4.5pt
    \small
    \caption{\small Trait Fidelity Evaluation using BioCAP captions as biologically informed external references.}
    \begin{tabular}{lcccc}
        \toprule
        Model & {BERT$\uparrow$} & {CLIP$\uparrow$} & {ROUGE-L$\uparrow$} & {Mistral-7B$\uparrow$} \\
        \midrule
        SD 1.5 (ZS)        & 0.28 & 0.80 & 0.24 & 0.88 \\
        SD 3.5-Large (ZS)  & 0.31 & 0.81 & 0.26 & 0.90 \\
        \flux (ZS)         & 0.34 & 0.83 & 0.29 & 0.91 \\
        SD 3.5-Large (LoRA) & 0.33 & \underline{0.87} & 0.28 & 0.91 \\
        \flux (LoRA)        & 0.31 & 0.85 & 0.28 & 0.91 \\
        \midrule
        \textbf{\ourmethod~(w/BioCLIP)}  & \textbf{0.41} & \textbf{0.88} & \textbf{0.32} & \textbf{0.93} \\
        \textbf{\ourmethod~(w/TaxaBind)} & \underline{0.36} & 0.86 & \underline{0.30} & \underline{0.92} \\
        \bottomrule
    \end{tabular}
    \vspace{-2mm}
    \label{tab:inat-caption-trait-fidelity}
\end{table}

\subsection{Free-form Text Conditioning}
\label{sup:free_form_text}
\begin{figure}[h]
    \centering
    \includegraphics[width=0.8\linewidth]{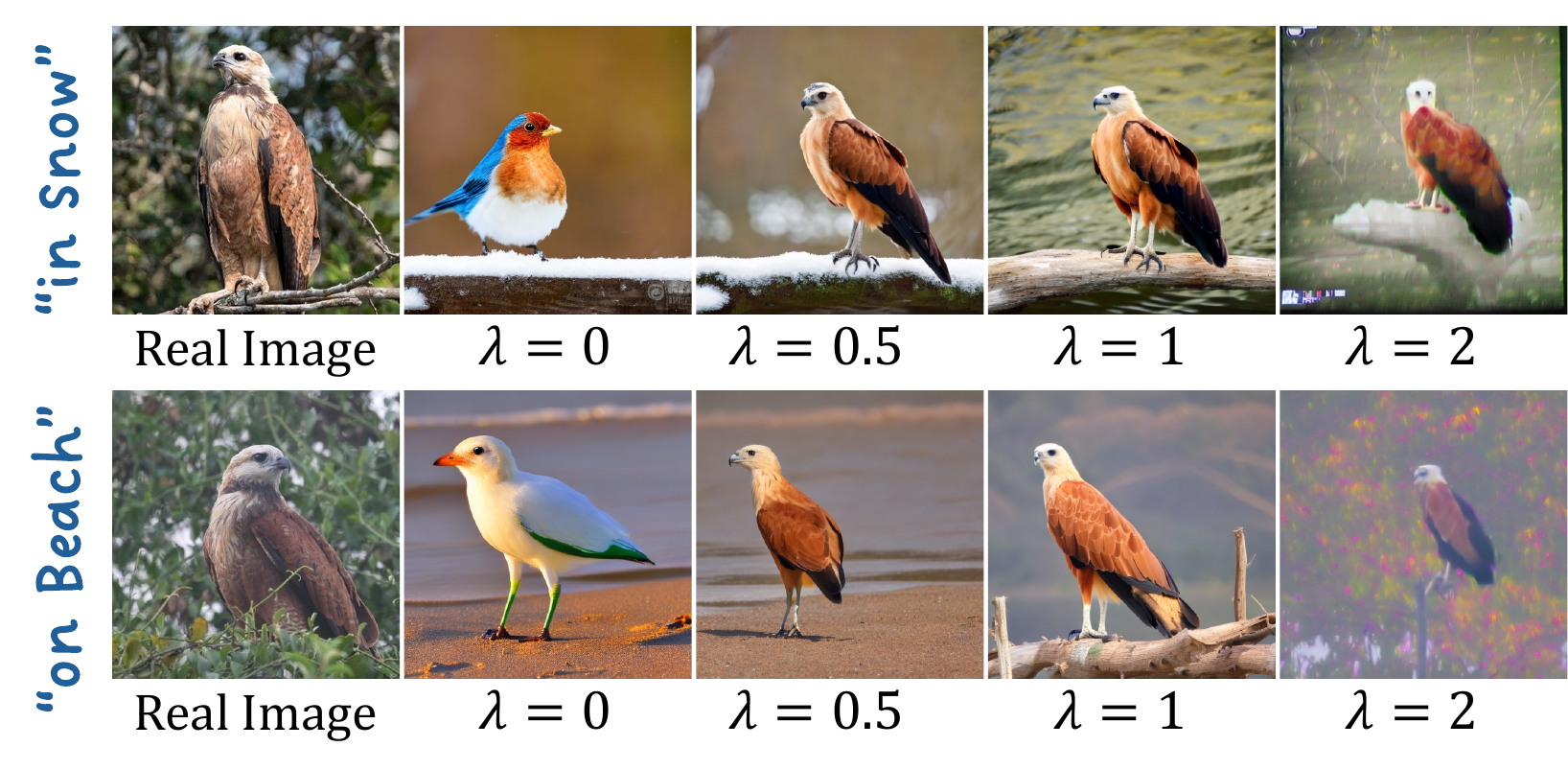}
    \vspace{-2mm}
    \caption{\small Different weighting factors $\lambda$ during inference for free-form text generation. Intermediate values (e.g., $\lambda = 0.5$) balance both conditions and yield morphology-faithful generations. Note that the setting of this figure is different from Figure~\ref{fig:ablation_merged} in the main paper, where we show general generation results.}
    \label{fig:lambda-supp}
\end{figure}

\begin{figure}[h]
    \centering
    \includegraphics[width=1\linewidth]{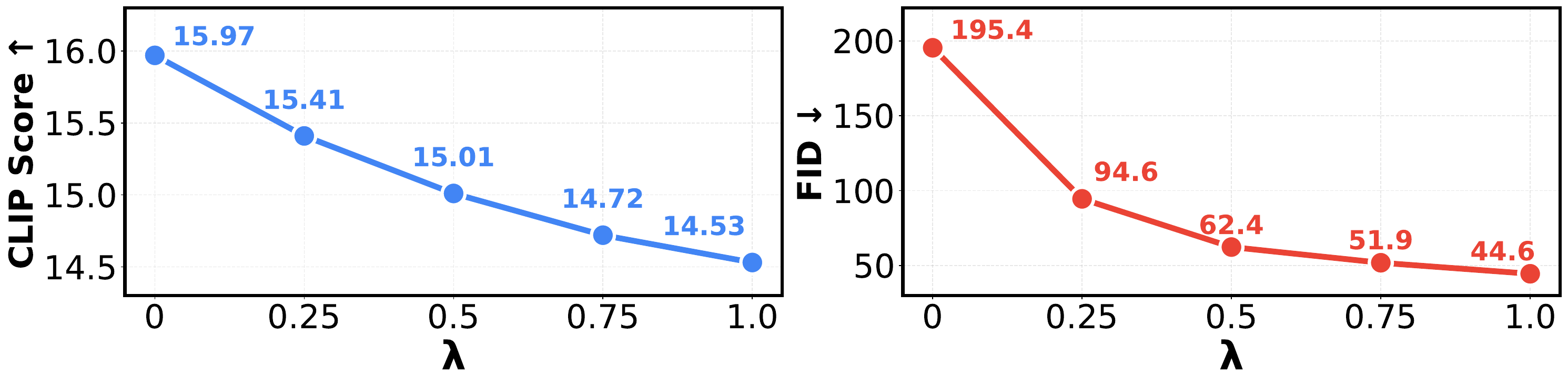}
    \vspace{-4mm}
    \caption{\small FID indicates species alignment; CLIP Score represents free-form text alignment (\wrt pose, background, etc.).}
    \label{fig:lambda-abblation}
    \vspace{-4.5mm}
\end{figure}

Figure~\ref{fig:lambda-supp} shows additional free-form prompting results enabled by \ourmethod. We set $\lambda=1$ during training (Eq.~\ref{eq:dual_attention}) and use $\lambda=1$ at inference for our default biologically-grounded generation setting, where the text branch is fixed to ``a photo of $y_{\text{taxa}}$'' and biological precision is the primary goal. We vary $\lambda$ \emph{only at inference} to enable free-form text control. With $\lambda=0.5$, the model maintains species-faithful morphology while supporting diverse contextual prompts. We do not tune $\lambda$ beyond these settings. Figure~\ref{fig:lambda-abblation} shows a quantitative analysis characterizing the effect of $\lambda$ on image quality and text alignment and motivates $\lambda=0.5$ for free-form prompting.

\subsection{Additional Qualitative Results}
\label{sup:additional_qualitative}
\begin{figure*}[ht!]
    \centering
    \includegraphics[width=1\linewidth]{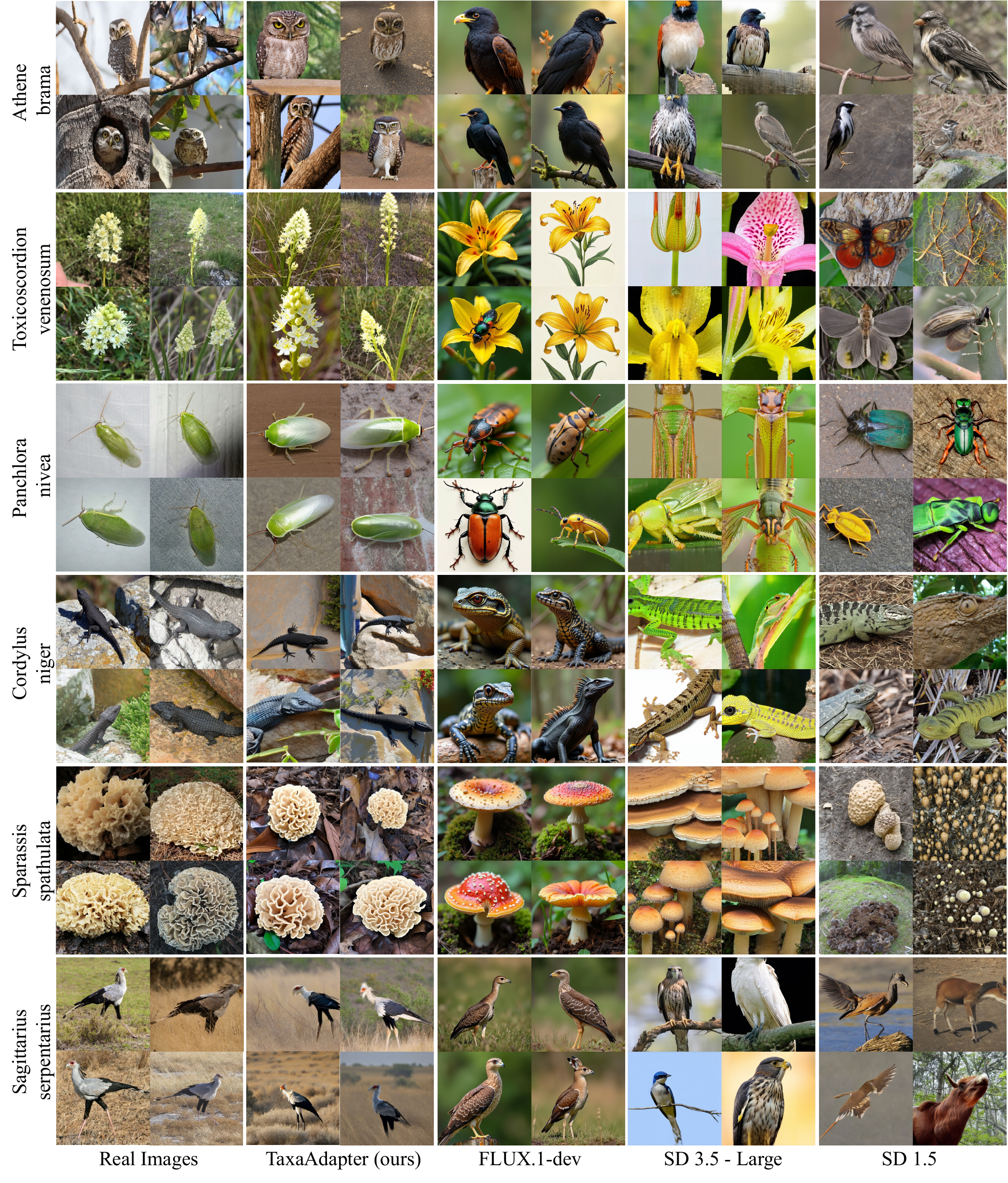}
    \caption{\small Additional qualitative results on \treeoflife-1M.}
    \label{fig:qualitative-1}
\end{figure*}

\begin{figure*}[ht!]
    \centering
    \includegraphics[width=1\linewidth]{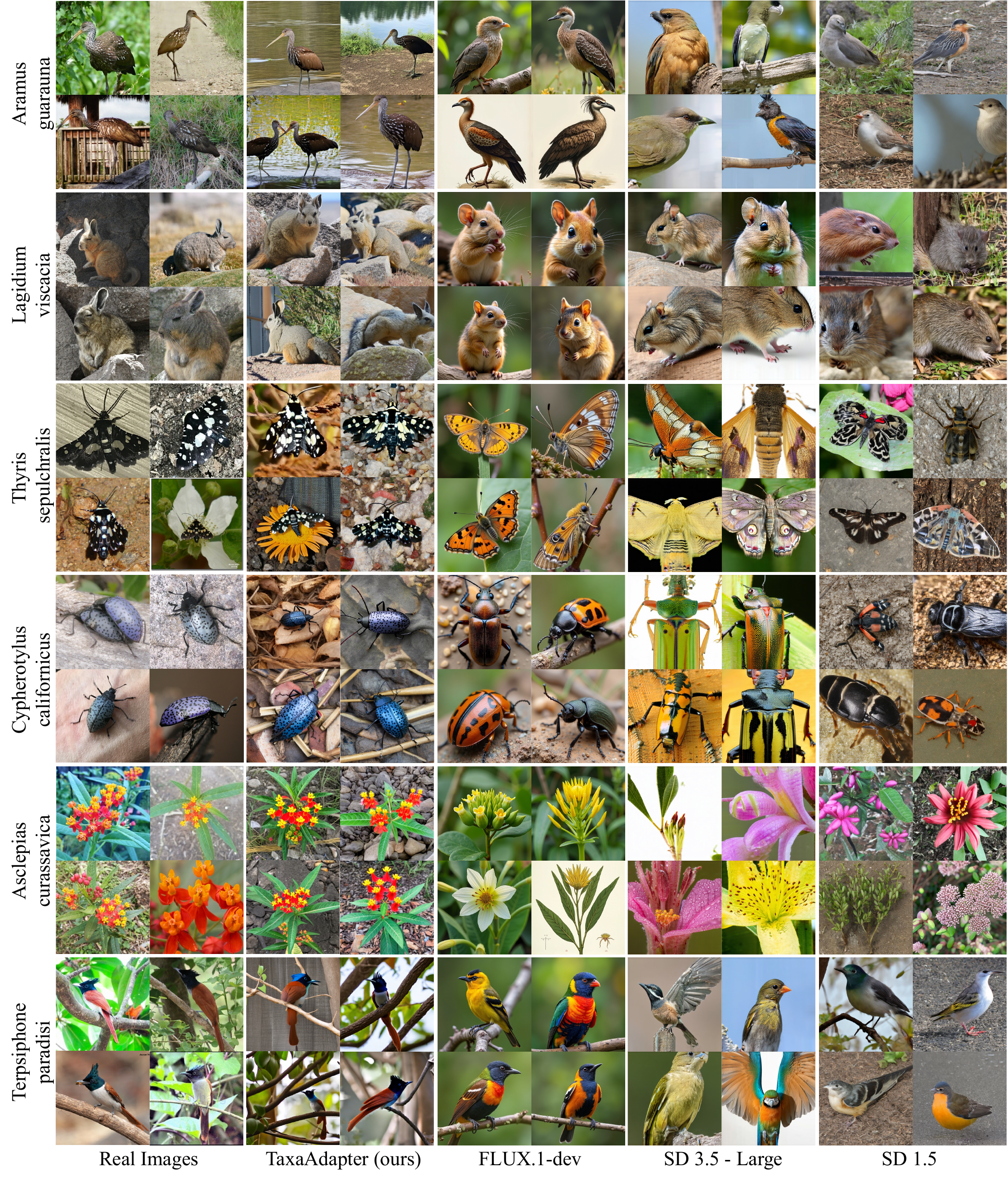}
    \caption{\small Additional qualitative results on iNaturalist.}
    \label{fig:qualitative-2}
\end{figure*}

\begin{figure*}[ht!]
    \centering
    \includegraphics[width=1\linewidth]{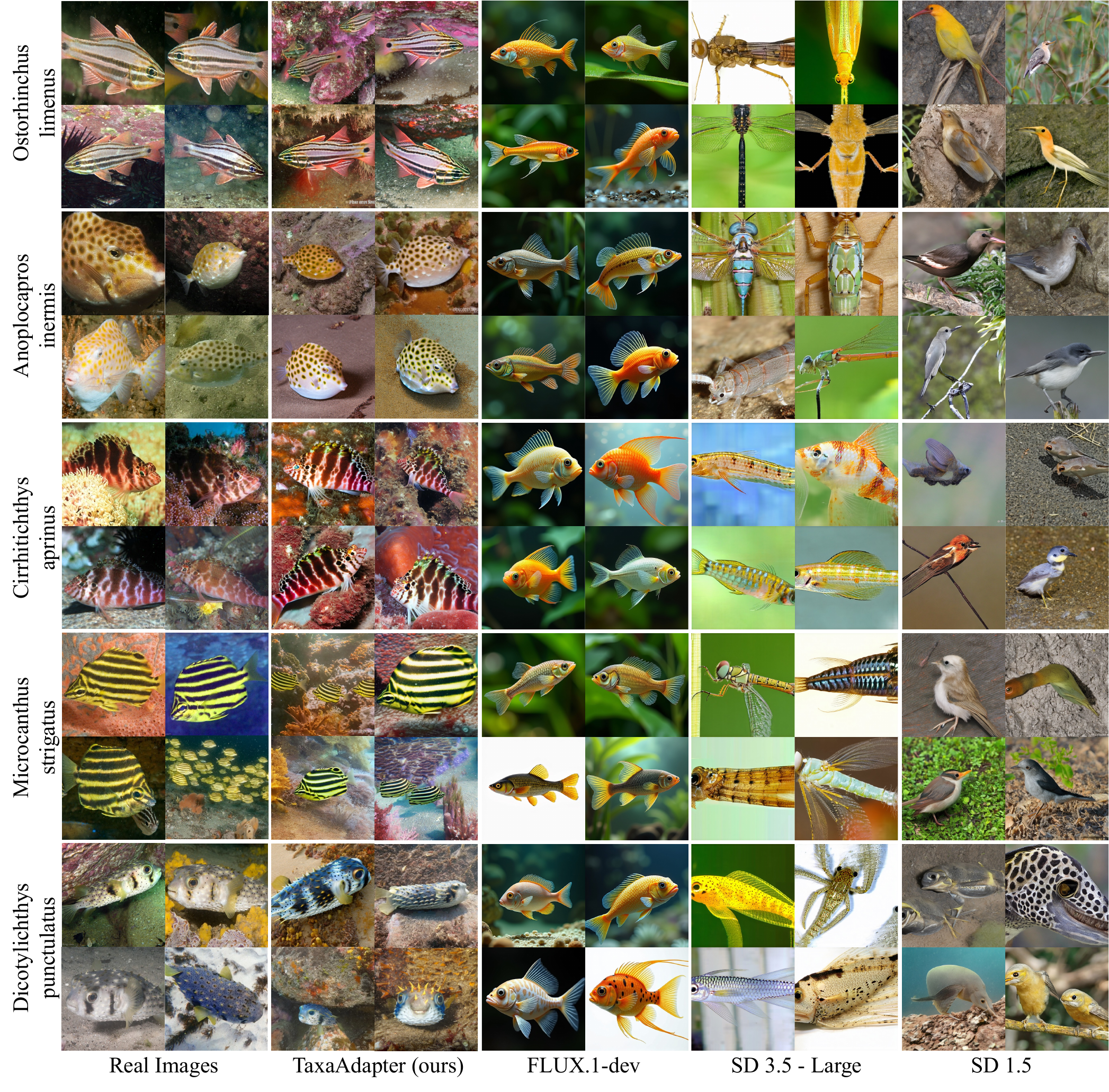}
    \caption{\small Additional qualitative results on FishNet.}
    \label{fig:qualitative-fishnet}
\end{figure*}

Figures~\ref{fig:qualitative-1}, \ref{fig:qualitative-2} and \ref{fig:qualitative-fishnet} show additional qualitative results of \ourmethod compared to the baselines.

\subsection{Additional OOD results}
\label{sup:additional_ood_results}
We additionally evaluate \ourmethod on 100 fish species from the FishNet~\cite{khan2023fishnet} dataset that were absent from both VTM pretraining and \ourmethod training, using the model trained only on the \treeoflife-1M dataset. Table~\ref{tab:ood_fishnet} shows that \ourmethod substantially improves species-level fidelity, achieving 8.30\% CAS@1 and 24.60\% CAS@5, compared with the strongest baseline results of 1.90\% and 10.70\%, respectively. It also achieves the best LPIPS (0.773), although SD3.5 obtains a better FID. These results extend the OOD evaluation beyond birds to a taxonomically distinct group and show a consistent improvement in species fidelity for previously unseen fish species.

\begin{table}[t]
    \centering
    \tabcolsep 4.5pt
    \small
    \caption{\small OOD results of TaxaAdapter on 100 FishNet species.}
    \label{tab:ood_fishnet}
    \begin{tabular}{lcccc}
        \toprule
        Model & FID $\downarrow$ & LPIPS $\downarrow$ & CAS@1 $\uparrow$ & CAS@5 $\uparrow$ \\
        \midrule
        SD1.5       & \underline{73.65} & \underline{0.799} & 1.80\% & 5.70\% \\
        SD3.5       & \textbf{64.83} & 0.811 & \underline{1.90\%} & \underline{10.70\%} \\
        FLUX.1-dev  & 85.32 & 0.802 & 1.30\% & 7.30\% \\
        \midrule
        \textbf{TaxaAdapter} & 82.48 & \textbf{0.773} & \textbf{8.30\%} & \textbf{24.60\%} \\
        \bottomrule
    \end{tabular}
\end{table}

For OOD on birds, as in Section~\ref{sec:emergent}, the 51 CUB test species are unseen during \ourmethod training for the model trained on the iNat-mini dataset. Among them, 16 species were present during VTM pretraining, whereas 35 were also absent from VTM pretraining after matching the taxonomic strings. The latter subset evaluates generalization to species unseen by both the adapter and the pretrained VTM, while the former evaluates whether \ourmethod can transfer knowledge already represented by the VTM to species excluded from adapter training, as shown in Tables \ref{tab:ood_cub_35} and \ref{tab:ood_cub_16}. We can further characterize the 35 species by their taxonomic proximity to the \ourmethod training distribution: 14 classes share the same genus, whereas the remaining 21 share a family but not a genus+species with their closest training relative.

\begin{table}[h]
    \centering
    \tabcolsep 4.5pt
    \small
    \caption{\small Results on 35 CUB species unseen during both VTM pretraining and \ourmethod training.}
    \label{tab:ood_cub_35}
    \begin{tabular}{lcccc}
        \toprule
        Model & FID $\downarrow$ & LPIPS $\downarrow$ & CAS@1 $\uparrow$ & CAS@5 $\uparrow$ \\
        \midrule
        SD1.5       & \textbf{37.03} & 0.792 & 7.14\% & 25.43\% \\
        SD3.5       & 45.24 & 0.794 & 4.29\% & 18.57\% \\
        FLUX.1-dev  & 57.71 & \textbf{0.769} & 3.43\% & 16.57\% \\
        \midrule
        \textbf{\ourmethod} & \underline{42.61} & \underline{0.788} & \textbf{29.43\%} & \textbf{60.57\%} \\
        \bottomrule
    \end{tabular}
\end{table}

\begin{table}[h]
    \centering
    \tabcolsep 4.5pt
    \small
    \caption{\small Results on 16 CUB species unseen during \ourmethod training only.}
    \label{tab:ood_cub_16}
    \begin{tabular}{lcccc}
        \toprule
        Model & FID $\downarrow$ & LPIPS $\downarrow$ & CAS@1 $\uparrow$ & CAS@5 $\uparrow$ \\
        \midrule
        SD1.5       & 82.28 & 0.783 & 15.00\% & 26.88\% \\
        SD3.5       & 80.21 & 0.806 & 16.88\% & 35.62\% \\
        FLUX.1-dev  & 90.16 & 0.816 & 13.75\% & 28.75\% \\
        \midrule
        \textbf{\ourmethod} & \textbf{73.23} & \textbf{0.753} & \textbf{59.38\%} & \textbf{85.00\%} \\
        \bottomrule
    \end{tabular}
\end{table}

\subsection{Mixed or Hybrid Species}
\begin{figure}[t]
    \centering
    \includegraphics[width=1\linewidth]{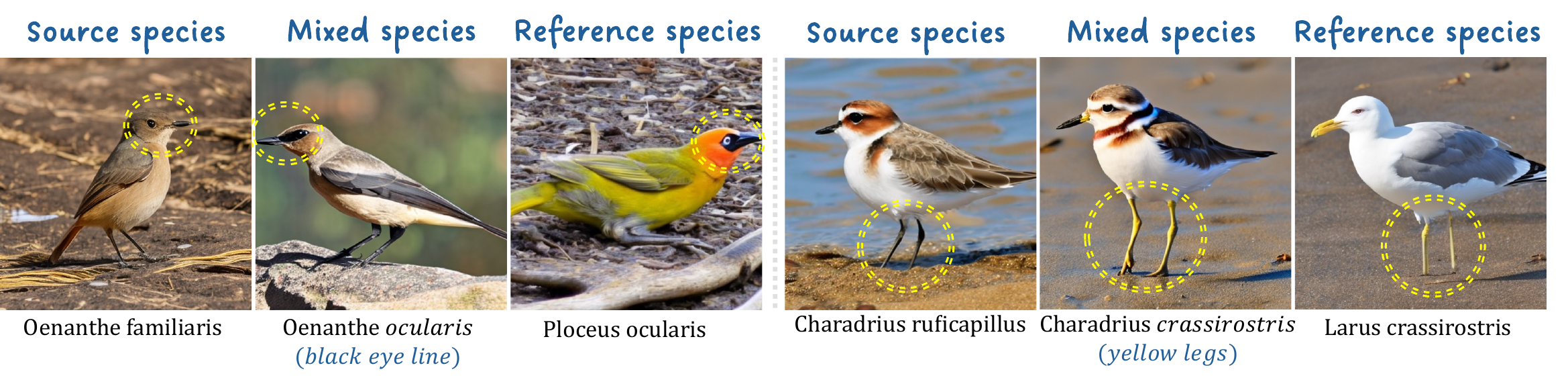}
    \vspace{-4mm}
    \caption{\small Synthetic ``mixed or hybrid'' species via taxonomic name mixing.}
    \label{fig:mixed_species}
    \vspace{-3mm}
\end{figure}

Beyond synthesizing known taxa, we show that \ourmethod exhibits compositional behavior in the VTM-conditioned embedding space. We do this by modifying the taxonomic name by mixing the names of two different species, for example, given ``Oenanthe familiaris'' and ``Ploceus ocularis'', we construct the hybrid species ``Oenanthe \textit{ocularis}'' and synthesize samples for this hybrid taxon. As shown in Figure~\ref{fig:mixed_species}, the mixed species inherits the black eye line trait from the reference species, demonstrating that VTM-aligned taxonomy embeddings can generate images that combine fine-grained morphological traits associated with different taxonomic levels. 
We, however, note that such mixed or hybrid species do not exist and are hence difficult to validate. We present them as an interesting direction for future studies for CRISPR-style~\cite{crispr} experiments for trait/gene editing.

\section{Additional Ablations and Analyses}
\label{sup:ablations}

\subsection{Compatibility with Different VTMs}
Tab.~\ref{tab:vtm_compatibility} shows the effect of varying the VTM backbone from BioCLIP to TaxaBind~\cite{sastry2025taxabind} and BioTrove-CLIP~\cite{yang2024biotrove}. 
We see that all VTM backbones achieve comparable gains over the CLIP baseline, demonstrating that \ourmethod works robustly with different VTMs and real benefits stem from domain-matched taxonomy encoders and not just fine-tuning.

\begin{table}[h]
\centering
\small
\captionof{table}{Analysis of taxa encoder in \ourmethod. Models are trained on iNat-mini-birds (1500 species) and evaluated on 650 bird species. 
}
\begin{tabular}{llccc}
\toprule
& {Taxa Encoder} & {FID}$\downarrow$ & {LPIPS}$\downarrow$ \\
\midrule
(a) & CLIP~\cite{radford2021learning} & 29.16 & 0.747 \\
(b) & BioTrove-CLIP~\cite{yang2024biotrove} & 21.51 & 0.745 \\
(c) & BioCLIP~\cite{gu2025bioclip} & 17.08 & 0.745 \\
(d) & TaxaBind~\cite{sastry2025taxabind} & 16.61 & 0.742 \\
\bottomrule
\end{tabular}
\label{tab:vtm_compatibility}
\end{table}

\subsection{Replacing CLIP with BioCLIP}
\label{sup:vtm_variants}
In Section~\ref{sec:ablations}, we studied the key design choices of \ourmethod. Here, we provide further justification for the taxonomy encoder design.

\mypara{Na\"ive fine-tuning with taxonomy (Table~\ref{tab:bioclip_replace_clip}-(b)).}
In the main paper, we describe that directly fine-tuning generative models with taxonomy information leads to suboptimal performance. Specifically, we take a subset of bird species from the iNat-mini dataset and train the model on 1500 species while evaluating on 650 species, sampling one species per \code{Genus} level in the taxonomy. We use zero-shot SD~1.5 as the baseline (row (a) in Table~\ref{tab:bioclip_replace_clip}). We then fine-tune SD~1.5 with LoRA by feeding taxonomy labels into its CLIP encoder. While this yields improvements in FID and LPIPS, it still underperforms \ourmethod.

\mypara{Replacing CLIP with BioCLIP (Table~\ref{tab:bioclip_replace_clip}-(c)).}
Next, we attempt the non-trivial substitution of the SD~1.5 CLIP text encoder with a VTM, BioCLIP (row (c)). Unlike the CLIP text encoder used in Stable Diffusion, which exposes a sequence of contextualized token embeddings of shape $[77 \times 768]$ for cross-attention, BioCLIP is typically used via its pooled and projected embeddings for classification tasks, having embeddings of shape $[1 \times 768]$. Directly feeding this pooled vector into the diffusion model would break the expected interface and discard the token-level structure. To make BioCLIP compatible with the diffusion backbone, we bypass its pooling and projection head and get the final-layer token embeddings to obtain the conditioning sequence for cross-attention of size $B \times 77 \times 768$. 

Despite these efforts, this variant achieves worse performance compared with naively fine-tuning the CLIP encoder.
This suggests that the model requires a significant amount of training to achieve taxonomy-to-image alignment. Also, since Stable Diffusion is largely pre-trained with CLIP, it is not easily replaceable without pre-training on a large dataset. All the fine-tuning was done similarly for 100 epochs using the same hyperparameters as \ourmethod.

\subsection{Compatibility with Other Diffusion Backbones}
\label{sup:sdxl}

One of the goals of \ourmethod\ is to serve as a plug-in taxonomy adapter that can be attached to different text-to-image backbones with dual-text encoders without fine-tuning of the model. To test this, we now train the model on SDXL~\cite{podell2023sdxl} for $512\times512$ images. In both cases, we use the original BioCLIP embeddings.
Table~\ref{tab:backbones} shows the results. We observe that upgrading the backbone from SD~1.5 to SDXL consistently improves both image quality and taxonomy–image alignment. FID drops from $17.08$ to $13.39$, while LPIPS remains essentially unchanged. These results indicate that the adapter cleanly factors out taxonomy control from low-level image synthesis: stronger backbones directly translate into better fine-grained generations, while the taxonomy branch continues to provide stable species-level guidance. Crucially, no additional tuning of the SDXL backbone is required beyond training the same lightweight adapter, demonstrating that \ourmethod\ is generalizable.

\begin{table}[h]
  \centering
  \small
  \tabcolsep 8pt
   \caption{Results of \ourmethod using different diffusion backbones on the iNat-mini birds subset. Both models use the same BioCLIP taxonomy encoder.}
    \begin{tabular}{lcc}
      \toprule
      {SD backbone} & {FID} $\downarrow$ & {LPIPS} $\downarrow$ \\
      \midrule
    TaxaAdapter (SD1.5)  & 17.08  & 0.745 \\
      TaxaAdapter (SDXL)    & 13.39  & 0.743 \\
      \bottomrule
    \end{tabular}
  \label{tab:backbones}
\end{table}

\subsection{Taxonomic Name vs. Common Name}
\label{sup:common_vs_taxonomic_name}
For state-of-the-art models like \flux, we don't know what data they were trained on. We evaluate its performance using both taxonomic names and the common names of the species for comparison. Concretely, we evaluate \flux\ on the \treeoflife-1M subset of 500 species using two prompt variants per species: (i) the full Linnaean taxonomic string (\emph{taxonomic name}, TN), for example, ``\emph{Animalia, Chordata, Aves, Passeriformes, Passerellidae, Spizella, breweri}'' and (ii) the common name (CN), for example,  ``Brewer's sparrow''. \ourmethod is evaluated with taxonomic-name prompts, which are also used as input to the VTM.

Table~\ref{tab:common-v-taxanomy} shows that \flux\ is much more aligned with common-name prompts. Switching from TN to CN reduces FID (from $90.09$ to $68.29$), and nearly doubles CAS@1/5. However, even under the favorable setting of using common names, \flux\ (CN) remains far behind our taxonomy-aware \ourmethod\ (TN). With taxonomic-name prompts only, \ourmethod attains an FID of $29.87$. These suggest that explicit taxonomy-aware conditioning is essential for robust species-level control in biodiversity synthesis.

\begin{table}[h]
\centering
\tabcolsep 3pt
\small
\caption{\small Quantitative comparison on \treeoflife-1M. Results are evaluated using 500 species on \treeoflife-1M. For \flux, we show the difference in performance when using the taxonomic name vs the common name.
TN: taxonomic name as prompt. CN: common name as prompt.}

\begin{tabular}{lcccc}
\toprule
Model & {FID$\downarrow$} & {LPIPS$\downarrow$} & {CAS@1$\uparrow$} & {CAS@5$\uparrow$} \\
\midrule
\flux (TN) & 90.09 & 0.800 & 12.44\% & 31.14\% \\
\flux (CN) & 68.29 & 0.803 &  21.28\% & 42.42\% \\
\midrule
\textbf{\ourmethod (TN)} & \textbf{29.87} & \textbf{0.734} & \textbf{61.98\%} & \textbf{82.96\%} \\
\bottomrule
\end{tabular}
\label{tab:common-v-taxanomy}
\end{table}

\subsection{T2I Baseline Using Detailed Morphological Descriptions / Retrieval-augmented Prompts}
\label{sup:biocap_t2i}
We use trait-descriptive prompts in text-to-image models as a useful complementary approach to generating morphologically accurate images when such trait descriptions are available. We conduct experiments using BioCAP \cite{zhang2025biocap} captions as text prompts, which contain detailed descriptions of species appearance, traits, and morphology. This setting also represents a retrieval-based approach where a detailed species description is retrieved (from BioCAP) and supplied to a strong text-to-image model.

We extend Table~\ref{tab:inat} results in Table~\ref{tab:biocap_t2i} and show that BioCAP prompting improves over the taxonomy-only baselines (Flux FID drops from 74.14 to 42.25), highlighting that richer morphological descriptions are beneficial for generation. However, these models still fall behind \ourmethod, particularly in species identity, where FLUX.1-dev achieves CAS@1 of 11.69\% vs 61.36\% for \ourmethod.

This shows that text captions certainly provide an opportunity to increase morphologically faithful generations (when available), but supplying descriptive text alone is hard to scale and also insufficient to match the benefits of the image-aligned taxonomic features learned directly from images without any trait labels. \ourmethod does not involve training and curation of any corpus of textual descriptions, making it trait label-free. A promising future direction is to explore combining descriptive text with VTM conditioning in \ourmethod.

\begin{table}[t]
    \centering
    \tabcolsep 4.5pt
    \small
    \caption{\small Quantitative comparison of TaxaAdapter with zero-shot (ZS) text-to-image generation models using BioCAP captions.}
    \label{tab:biocap_t2i}
    \begin{tabular}{lcccc}
        \toprule
        Model & FID $\downarrow$ & LPIPS $\downarrow$ & CAS@1 $\uparrow$ & CAS@5 $\uparrow$ \\
        \midrule
        {SD 1.5 (ZS w/ taxonomic name)}        
        & 31.43 & 0.768 & 1.24\% & 2.86\% \\
        
        {SD 3.5-Large (ZS w/ taxonomic name)} 
        & 70.63 & 0.806 & 1.86\% & 4.48\% \\
        
        {\flux (ZS w/ taxonomic name)}                
        & 74.14 & 0.801 & 1.78\% & 4.21\% \\
        \midrule
        
        SD1.5 (ZS w/ BioCAP captions)       & 19.84 & 0.761 & 8.45\%  & 18.23\% \\
        SD3.5 (ZS w/ BioCAP captions)       & 36.85 & 0.783 & 13.23\% & 27.19\% \\
        FLUX.1-dev (ZS w/ BioCAP captions)  & 42.25 & 0.778 & 11.69\% & 25.04\% \\
        \midrule
        \textbf{\ourmethod (w/ BioCLIP)}      & \textbf{13.31} & \textbf{0.742} & \textbf{61.36\%} & \textbf{81.99\%} \\
        \bottomrule
    \end{tabular}
\end{table}

\subsection{Isolating the Contribution of Hierarchical Taxonomy}

To isolate the contribution of hierarchical taxonomic conditioning, we train \ourmethod models on 1103 iNaturalist species, selecting one species from each family, and altering the taxonomic string supplied to the VTM. Specifically, we compare the canonical seven-level taxonomy against two controls: (i) randomly permuting the same taxonomic ranks, thereby disrupting their canonical root-to-leaf order, and (ii) replacing the higher-level ranks with those of a different species, thereby producing a biologically inconsistent hierarchy.

Table~\ref{tab:taxonomy_ordering} shows that the complete, correctly ordered seven-level taxonomy achieves the highest species-identity performance, with 73.64\% CAS@1 and 86.70\% CAS@5. Randomizing the order produces the worst results, reducing CAS@1 to 43.44\% and degrading FID from 13.12 to 16.87. More importantly, corrupting the higher-level ranks while retaining the lower-level identity also decreases CAS@1 to 71.11\% and CAS@5 to 84.73\%. This indicates that biologically consistent higher ranks provide complementary lineage context for preserving fine-grained species morphology.

BioCLIP was pretrained using a mixture of scientific names (genus+species), common names, and complete taxonomic strings. Hence, sending a scientific name (genus+species) as input to BioCLIP is bound to implicitly recover the complete hierarchical information through the pretrained VTM. Hence, masking higher-level taxonomic information in the inputs to VTMs may not help us to evaluate the importance of higher levels of the taxonomy hierarchy.

\begin{table}[h]
    \centering
    \tabcolsep 4.5pt
    \small
    \caption{\small Ablation study on the effect of hierarchical taxonomic information in \ourmethod.}
    \label{tab:taxonomy_ordering}
    \begin{tabular}{lcccc}
        \toprule
        Hierarchical Taxonomy & FID $\downarrow$ & LPIPS $\downarrow$ & CAS@1 $\uparrow$ & CAS@5 $\uparrow$ \\
        \midrule
        Random order            & 16.87 & 0.748 & 43.44\% & 62.03\% \\
        Corrupted higher levels & 13.44 & 0.747 & 71.11\% & 84.73\% \\
        \midrule
        \textbf{All 7 levels}   & \textbf{13.12} & \textbf{0.743} & \textbf{73.64\%} & \textbf{86.70\%} \\
        \bottomrule
    \end{tabular}
\end{table}

\section{Limitations}
\label{sup:limitations}

While \ourmethod\ demonstrates strong taxonomy-aware generation across multiple biodiversity datasets, it has certain limitations.
One limitation of the current setup is that the free-form text conditioning is biased toward backgrounds and contexts that appear in the training data. When prompts describe species in rare or unseen environments (e.g., ``a photo of \textit{Spizella breweri} perched on the roof of a car''), the model often fails to generate accurate images.

Another limitation is that our model still inherits challenges from data quality and sparsity. Species or clades that occupy only a small fraction of the frame in the images remain challenging. In such cases, we observe failures where the generated samples blur fine-grained markings or allow background textures to dominate.

Lastly, our evaluation of visual trait fidelity relies on large vision–language models and LLM-based trait summarization. While these provide a scalable proxy in the absence of expert-annotated trait captions, they introduce their own biases and may under- or over-emphasize certain traits relative to expert-defined descriptors. A more thorough integration with expert-curated trait databases and human evaluations is an important direction for future work.

\section{Ethics and Social Impact}
\label{sec:ethics}
Our work focuses on species generation using public image datasets and taxonomy data, with the aim of accelerating scientific discovery, research, and usage. We do not use any human subjects or personally identifiable information, and we do not foresee any negative impacts or ethical concerns with our work. The primary intended use of \ourmethod\ is to support scientific exploration, education, and data augmentation for biodiversity and ecological research.

\end{document}